\documentclass[10pt,twocolumn,letterpaper]{article}

\usepackage[pagenumbers]{cvpr} 

\usepackage{microtype}
\usepackage{graphicx}
\usepackage{booktabs} 
\usepackage{algorithm}
\usepackage{algorithmic}
\usepackage{amsmath,amsfonts,amssymb,amsthm} 
\usepackage{mathtools}

\usepackage{float}
\usepackage{adjustbox}
\usepackage{makecell} 
\usepackage{csquotes}

\usepackage{pifont}

\usepackage{listings}
\usepackage{xcolor}
\usepackage[table]{xcolor}

\usepackage{mathtools}

\newtheorem{lemma}{Lemma}

\newtheorem{proposition}{Proposition}

\usepackage{etoc}

\usepackage{titletoc}
\usepackage{tcolorbox}

\usepackage{booktabs}
\usepackage{arydshln}
\usepackage{multirow}
\usepackage{xcolor}
\usepackage[table]{xcolor}
\usepackage{threeparttable}
\usepackage{url}

\newcommand{\reconpanel}[2]{%
    \begin{subfigure}{\textwidth}
        \centering
        \includegraphics[
            width=\linewidth
        ]{#1}
        \caption{}\label{#2}
    \end{subfigure}\par
}

\newtheorem*{restatedproposition1}
    {Proposition~\ref{prop:flow_y_stability}}
\newtheorem*{restatedproposition2}
    {Proposition~\ref{prop:conditional_stability}}

\AtBeginDocument{\etocdepthtag.toc{main}} 

\definecolor{cvprblue}{rgb}{0.21,0.49,0.74}
\usepackage[pagebackref,breaklinks,colorlinks,allcolors=cvprblue]{hyperref}

\def\paperID{26973} 
\def\confName{CVPR}
\def\confYear{2026}

\title{Lipschitz-Controlled Flow Matching for Robust Image Restoration}

\author{
    Sedjro Salomon Hotegni$^{1,2}$ \qquad
    Konstantin Sonntag$^{1}$ \qquad
    Paul-Christian Bürkner$^{2}$ \qquad Sebastian Peitz$^{1}$ \\
    $^{1}$Paderborn University \qquad
    $^{2}$TU Dortmund University \\
    {\tt\small
    \{sedjro.salomon.hotegni, 
konstantin.sonntag,  sebastian.peitz\}@uni-paderborn.de} \\
    {\tt\small paul.buerkner@gmail.com}
}

\begin{document}
\maketitle
\begin{abstract}
Flow-matching models can produce high-quality image restorations, but their learned dynamics may amplify small input perturbations and make reconstructions unstable. We introduce
\textit{LipsFlower} to reduce this amplification. We derive
stability bounds for Flow-from-$Y$, which initializes the ordinary
differential equation from a noise-perturbed observation, as well as for
observation-conditioned posterior sampling, which conditions the
velocity field on the observation throughout integration.
Based on this analysis, we develop LipHDiT, a Lipschitz-aware
hourglass diffusion transformer that serves as the vector-field architecture of LipsFlower. LipHDiT incorporates stabilized approximate spectral normalization, bounded modulation, and convex residual mixing. We also use adversarial finite-radius gain regularization
at sampled interpolation states to penalize excessive sensitivity
to the state and, for conditional sampling, the observation.
We test both formulations on blind face restoration, measuring adversarial robustness and the perceptual quality of the reconstructions. LipsFlower reduces perturbation amplification while keeping restoration quality
competitive with the evaluated baselines. These results suggest that Lipschitz-aware flow matching can improve the stability of generative image restoration.
\end{abstract}    
\section{Introduction}
\label{sec:introduction}

Image restoration seeks to recover clean images
from degraded observations. One particularly challenging case is
blind face restoration, which aims to recover a
high-quality face image from an observation affected by unknown
degradations, such as blur, noise, downsampling, and compression.
The problem is ill-posed: several clean faces may explain the
same degraded input. The output must therefore look realistic,
preserve facial identity and geometry, and stay consistent with
the observation. Recent generative methods use face priors to
synthesize plausible
details~\cite{wang2021towards,gu2022vqfr,zhou2022towards,yue2024difface}.
But a convincing output can still be sensitive to small input
perturbations. 
\begin{figure}[htbp]
    \centering
    \includegraphics[width=\linewidth]
        {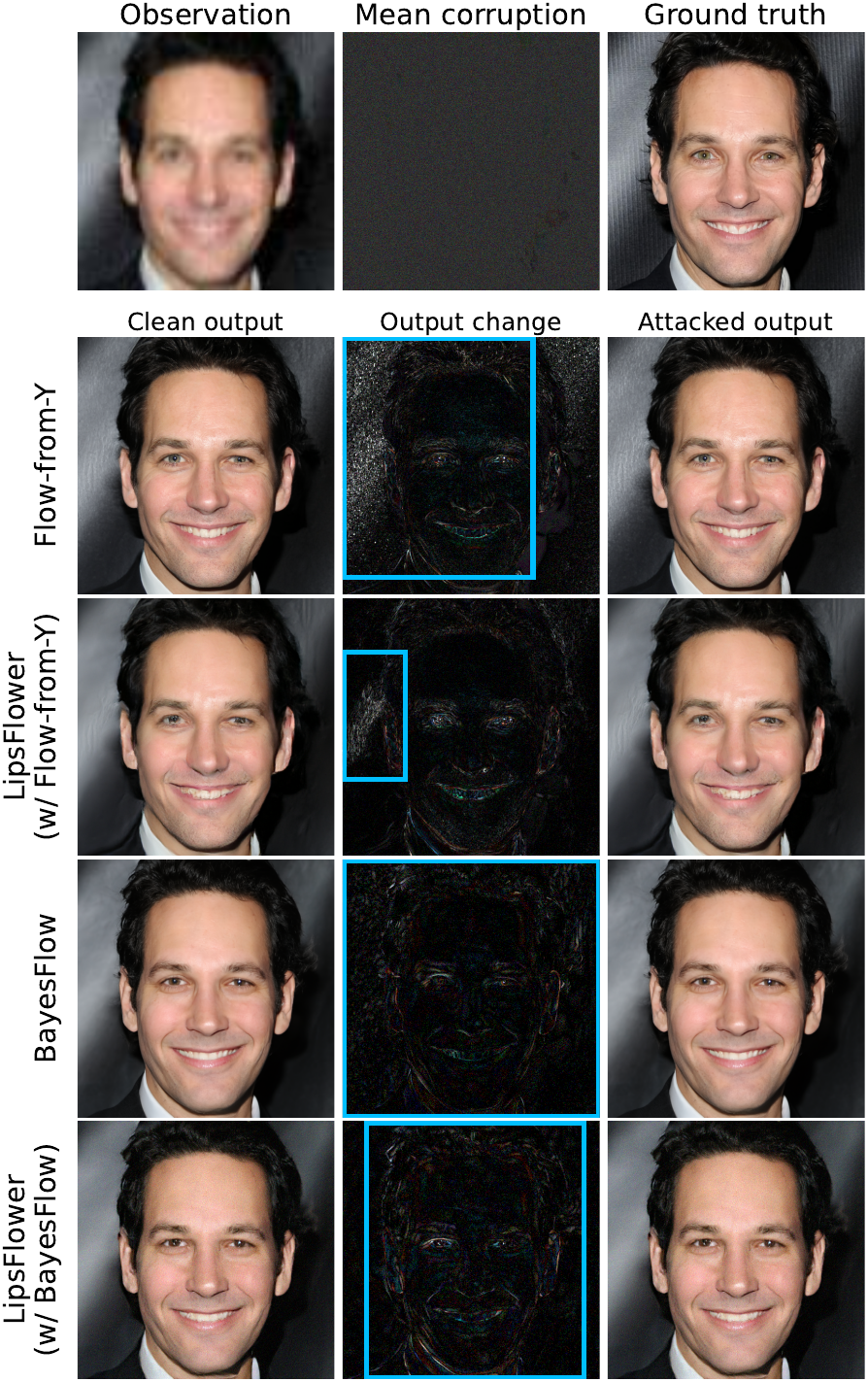}
        \caption{\textbf{Robustness to observation perturbations.}
        LipsFlower reduces output changes for both backbones under model-specific black-box SPSA attacks, with a larger reduction for Flow-from-$Y$.
        The mean perturbation and output differences are amplified for visibility.
        Blue boxes mark the largest above-threshold change regions,
        with thresholds matched within each backbone pair.}
    \label{fig:teaser}
    \vspace{-0.75cm}
\end{figure}
For human faces, even a slight change in the input
can cause an image restoration model to produce noticeably different facial features, potentially
changing the person's apparent identity.
Metrics such as PSNR~\cite{huynh2008scope},
LPIPS~\cite{zhang2018unreasonable}, and FID~\cite{heusel2017gans}
assess reconstruction or distributional quality, but do not directly
measure input sensitivity. 
A model can score well on these metrics
and still produce very different reconstructions from nearly
identical observations.

Flow matching~\cite{lipman2023flow} learns a time-dependent
velocity field whose ordinary differential equation (ODE) transports
samples from a source distribution toward clean images. A common form of flow matching is rectified flow~\cite{liu2022flow},
in which models are trained on straight-line interpolations
between source and target samples. Recent work shows that these models can produce
visually convincing restorations~\cite{liu2022flow,ohayon2024posterior}. We consider two
formulations: \emph{Flow-from-$Y$}~\cite{ohayon2024posterior}, which
starts the ODE from a noise-perturbed degraded observation, and
\emph{observation-conditioned posterior sampling}~\cite{zhu2024flowie}, which starts from a simple base distribution (e.g., multivariate Gaussian) and conditions the velocity field on the observation throughout integration.
Neither formulation by itself guarantees that nearby observations
produce nearby reconstructions. In Flow-from-$Y$, an input
perturbation changes the initial state and propagates through the
dynamics. In conditional sampling, perturbing the observation directly
changes the velocity field. Our stability analysis in~\cref{sec:stability_analysis} relates these effects to the field's Lipschitz behavior: the velocity field's sensitivity to the evolving ODE state matters in both
formulations, while conditional sampling additionally
requires controlling the velocity field's sensitivity
to the observation. Standard flow-matching regression does not directly constrain the field's Lipschitz constants with respect to the ODE state or the observation.
Our goal is to reduce excessive amplification while preserving the capacity needed for realistic restoration. This motivates controlling selected sources of
amplification rather than imposing an excessively small global Lipschitz constant. The latter may restrict the nonlinear transformations needed to restore heavily degraded images and recover facial detail.

We introduce \emph{LipsFlower}, a flow-matching framework that
combines architectural controls with adversarial finite-radius gain
regularization to reduce input sensitivity. Its velocity network,
\emph{LipHDiT}, builds on the hourglass diffusion transformer
(HDiT)~\cite{crowson2024scalable}. It uses stabilized approximate
spectral normalization, controlled normalization and time
modulation, and convex residual and skip mixing. Our architectural contribution lies in adapting
and integrating these controls within HDiT to improve restoration
stability. The conditional
variant of LipHDiT uses separate state and observation projection
stems with controlled fusion, targeting
the additional observation sensitivity identified by our analysis.
We also adapt adversarial Lipschitz
regularization~\cite{terjek2019adversarial} to velocity fields at
sampled flow-matching interpolation states. To do this, we search for local
state perturbations at several magnitude scales and use a separate
observation perturbation search to target conditional sensitivity.
A hinge penalty applies only when the finite-difference velocity
gain exceeds a prescribed threshold. Sampling across training
times extends regularization beyond the source and endpoint
without requiring full ODE rollouts. Together, these mechanisms
encourage empirical stability.
We evaluate both formulations on blind face restoration,
measuring reconstruction changes under adversarial input
perturbations, along with reconstruction and perceptual quality.
Our experiments show that architectural controls and gain
regularization reduce perturbation amplification while keeping
reconstruction fidelity and perceptual quality competitive with the evaluated baselines.
Our main contributions are:
\begin{itemize}
    \item We derive stability bounds for Flow-from-$Y$ and
    observation-conditioned posterior sampling, identifying the state and observation Lipschitz quantities that govern input
    sensitivity.

    \item We introduce LipsFlower, which combines a Lipschitz-aware
    hourglass transformer with adversarial finite-radius
    regularization of state sensitivity and, for conditional
    sampling, observation sensitivity.

    \item We evaluate both formulations on blind face restoration,
    assessing restoration-map
    sensitivity and reconstruction quality. The results
    show that LipsFlower improves robustness to observation
    perturbations, particularly in the Flow-from-$Y$
    formulation, while maintaining competitive
    reconstruction quality.
\end{itemize}

\section{Related Work}
\label{sec:related_work}

\paragraph{Blind face image restoration.}
Early deep-learning methods for blind face restoration (BFR) used facial geometry, semantic parsing maps, or component dictionaries to reduce ambiguity
\cite{li2018learning,chen2018fsrnet,li2020blind}. More recent generative
methods use learned priors to synthesize realistic facial details.
GPEN~\cite{yang2021gan} and GFP-GAN~\cite{wang2021towards} incorporate
generative face priors into restoration networks.
VQFR~\cite{gu2022vqfr}, RestoreFormer~\cite{wang2022restoreformer}, and
CodeFormer~\cite{zhou2022towards} use vector-quantized codebooks or
transformer-based retrieval of high-quality facial features.
These methods improve perceptual quality, but strong priors can also
generate details that the observation does not clearly support.
Diffusion models~\cite{ho2020denoising} have since been applied to BFR
through conditional generation, degradation removal, and identity-aware
guidance. DifFace~\cite{yue2024difface} uses a pretrained diffusion prior
for posterior estimation, while DR2~\cite{wang2023dr2} separates
degradation removal from content enhancement.
Region-adaptive conditioning~\cite{suin2024diffuse},
frequency-domain restoration~\cite{miao2024waveface}, and one-step
diffusion models~\cite{wang2025osdface} further improve identity
preservation, efficiency, or the perception--distortion trade-off.
Evaluation still focuses mainly on the quality of restored images.
Their consistency under small changes to the degraded observation
has received less attention. Standard distortion and perceptual
metrics do not directly show whether nearby observations produce
nearby reconstructions. Our experiments show that
the evaluated baselines can amplify small observation perturbations
into larger changes in their reconstructions.

\paragraph{Flow matching for image restoration.}
Flow-based restoration methods differ in how they use the degraded
observation. Plug-and-play methods such as
PnP-Flow~\cite{martin2025pnp} reuse a pretrained flow model
as an image prior within an iterative reconstruction algorithm
that promotes consistency with the degraded observation. Other methods transport degraded observations or preliminary
estimates toward the clean-image
distribution~\cite{ohayon2025posterior,fan2026ir}. For example, ELIR~\cite{cohen2025efficient} uses latent consistency flow matching to transport preliminary latent estimates with added Gaussian noise toward clean-image representations.
FlowIE~\cite{zhu2024flowie} instead learns a conditional rectified flow from Gaussian noise to clean-image representations, using initial restorations of the degraded inputs as guidance. 
PMRF~\cite{ohayon2025posterior} is closely related to our setting.
It transports an estimate of the posterior mean toward the clean-image
distribution, aiming for a better perception--distortion trade-off.
Its controlled study compares Flow-from-$Y$ with two conditional
rectified-flow baselines whose velocity fields are conditioned on
either the degraded observation or its posterior-mean estimate. Unlike PMRF's two-stage pipeline, our formulations
do not require a separate posterior-mean predictor.
Other recent methods use rectified flows to connect discriminative
and generative restoration or progressively refine images across
degradation levels~\cite{fan2026ir}. These methods focus mainly on
restoration quality, sampling efficiency, and transport design.
We instead study how perturbations to the degraded observation propagate
through the learned dynamics. We consider two pathways: the initial
condition in Flow-from-$Y$ and the observation supplied to the velocity
field in observation-conditioned posterior sampling.

\paragraph{Lipschitz control and robustness of learned dynamics.}
Methods for controlling input sensitivity in neural networks include
spectral normalization of linear operators~\cite{miyato2018spectral},
gradient penalties~\cite{gulrajani2017improved}, orthogonal or
norm-constrained parameterizations~\cite{cisse2017parseval}, and non-expansive activations~\cite{anil2019sorting}. GroupSort networks
combine norm-constrained linear maps with gradient-norm-preserving
activations to retain expressivity under Lipschitz constraints
\cite{anil2019sorting,tanielian2021approximating}.
Standard dot-product self-attention is not globally Lipschitz on
an unbounded domain, which has motivated normalized and otherwise
controlled attention mechanisms
\cite{kim2021lipschitz,pmlr-v139-dasoulas21a}.
For neural ODEs, the regularity of the vector field governs the
existence, uniqueness, and sensitivity of trajectories
\cite{NEURIPS2018_69386f6b}. This connection has been used to study
robustness, stability, and generalization in continuous-depth models
\cite{YAN2020On,NEURIPS2021_7d5430cf,NEURIPS2023_98ed250b}.
Recent work on plug-and-play flow restoration also links vector-field
Lipschitz regularity to numerical restoration error
\cite{jia2025plugandplayimagerestorationflow}.
Our analysis focuses on how perturbations to the degraded observation
affect the restoration map, both for flows initialized
from the observation and for flows conditioned on it.
\textit{LipsFlower} uses this analysis to guide two forms of control: architectural control and gain regularization.
First, LipHDiT limits amplification within the hourglass diffusion
transformer through controlled normalization and conditioning,
regularized linear projections, and scaled residual branches.
Second, since layer-wise global bounds can be
conservative and overly restrictive, we also regularize adversarial
finite-radius gain at sampled flow-matching interpolation states. This penalizes excessive local amplification across training times without imposing an overly small global Lipschitz constant on the whole model.
TVM~\cite{zhou2026terminal} is the closest work architecturally.
It identifies the lack of Lipschitz continuity in standard diffusion
transformers as a source of training instability and introduces
semi-Lipschitz modifications based on normalized attention,
normalization, modulation, and initialization.
Under a Lipschitz assumption, it relates a terminal-velocity
objective to distributional error in one- and few-step image
generation. This result concerns the distance between generated and target image distributions, not the sensitivity of individual reconstructions to their degraded observations.
We bound how observation
perturbations affect the final reconstruction when introduced
through the initial state or the observation condition.
We also regularize finite-radius amplification
at sampled interpolation states.
The goal is to make small observation changes
produce correspondingly small reconstruction changes, with
sampling randomness held fixed.









\section{Background and Problem Formulation}
\label{sec:problem}

\subsection{Blind Face Restoration}
\label{sec:bfr_problem}

Let \(X \in \mathcal{X}\subseteq\mathbb{R}^{d}\) denote a clean face image and
\(Y\in\mathcal{Y}\subseteq\mathbb{R}^{d}\) its degraded observation\footnote{The shared dimensionality
reflects the image-restoration setting considered here.
Conditional posterior sampling also supports observations in
\(\mathbb{R}^{m}\), with \(m\neq d\). More general observation structures would require a suitable observation-space metric and an adapted conditioning encoder whose sensitivity is also controlled.
We leave such architectural extensions to future work.}. During training, paired samples
\((X,Y)\sim p_{X,Y}\) are available, with \(Y\) generated by a stochastic
degradation process that may combine blur, downsampling, noise, and compression.
At inference time, the degradation parameters are unknown~\cite{wang2021towards, yue2024difface}. The objective is to
construct a potentially stochastic restoration map $\widehat{X}=\mathcal{R}(Y).$
Besides producing perceptually realistic and observation-consistent faces, we require
\(\mathcal{R}\) to be stable: \textit{a small perturbation of the degraded observation
    should not cause a disproportionately large change in the reconstruction when
    the model randomness is held fixed.}
This requirement is distinct from conventional restoration quality. Metrics
such as PSNR and LPIPS assess the relation between a reconstruction and its
reference, whereas stability concerns the variation of the restoration map
with respect to its input. We therefore study the response of
\(\mathcal{R}(Y)\) to nearby observations \(Y'=Y+\delta\), where
\(\delta\) may represent incidental corruption or an adversarially selected
perturbation\footnote{This notation also covers
transformations \(Y'=f(Y)\) in the same observation space by setting
\(\delta=f(Y)-Y\). Under the assumptions stated
in~\cref{sec:stability_analysis}, the stability bounds depend on
\(\|f(Y)-Y\|_2\), regardless of how the perturbation is obtained.}.

\subsection{Flow Matching}
\label{sec:flow_matching_background}

Let
\(Z_0\) be a source sample and let \(Z_1=X\) be a clean target image. We use the
linear probability path~\cite{lipman2023flow, liu2022flow}
\begin{equation}
    Z_t=(1-t)Z_0+tX,
    \qquad t\in[0,1],
    \label{eq:linear_probability_path}
\end{equation}
whose sample-wise target velocity is
\begin{equation}
    U_t=\frac{\mathrm{d}Z_t}{\mathrm{d}t}=X-Z_0.
    \label{eq:target_velocity}
\end{equation}

A neural velocity field \(v_\theta\) is trained by regressing this target:
\begin{equation}
    \mathcal{L}_{\mathrm{FM}}(\theta)
    =
    \mathbb{E}
    \left[
        \left\|
            v_\theta(Z_t,t)-\left(X-Z_0\right)
        \right\|_2^2
    \right],
    \label{eq:unconditional_fm_loss}
\end{equation}
where \(t\sim\mathcal{U}[0,1]\). At the population optimum, the learned field
approximates the conditional mean velocity $v^*(z,t) = \mathbb{E}\!\left[X-Z_0\mid Z_t=z\right].$
To generate a sample, we draw \(Z_0\) and integrate
\begin{equation}
    \frac{\mathrm{d}z_t}{\mathrm{d}t}
    =v_\theta(z_t,t),
    \qquad t\in[0,1].
    \label{eq:unconditional_flow_ode}
\end{equation}

We denote the flow map from time \(s\) to \(t\) by
\(\Phi^\theta_{s\rightarrow t}\), so the terminal sample is
\(\widehat{X}=\Phi^\theta_{0\rightarrow1}(Z_0)\). In practice,
we approximate this continuous-time flow with a numerical solver
using a finite number of steps.

This general construction can
be made observation-dependent in two ways: the degraded observation can determine the source state or condition the velocity field.
We consider both cases below.

\subsection{Observation-Dependent Restoration Flows}
\label{sec:observation_dependent_flows}

\subsubsection{Flow-from-Y}
\label{sec:flow_from_y}

Flow-from-$Y$ starts from a noisy version of the degraded
observation~\cite{ohayon2025posterior}.
Let \(Y\) be the degraded observation in the same image space
as the clean image \(X\). The source state is
\begin{equation}
    Z_0^{Y}=Y+\sigma_s\epsilon,
    \qquad
    \epsilon\sim\mathcal{N}(0,I),
    \label{eq:flow_y_source}
\end{equation}
where \(\sigma_s\) controls the source-noise level. The training
path and target velocity are
\begin{equation}
    Z_t^{Y}=(1-t)Z_0^{Y}+tX,
    \qquad
    U_t^{Y}=X-Z_0^{Y}.
    \label{eq:flow_y_path}
\end{equation}

The velocity field depends on \(Y\) through the evolving state,
without receiving it as a separate input. Its flow-matching loss is
\begin{equation}
    \mathcal{L}_{\mathrm{F}Y}(\theta)
    =
    \mathbb{E}
    \left[
        \left\|
            v_\theta(Z_t^{Y},t)-U_t^{Y}
        \right\|_2^2
    \right].
    \label{eq:flow_y_loss}
\end{equation}

At inference, the restoration map is
\begin{equation}
    \mathcal{R}_{\mathrm{F}Y}(y;\epsilon)
    =
    \Phi^\theta_{0\rightarrow1}
    \left(y+\sigma_s\epsilon\right).
    \label{eq:flow_y_restoration}
\end{equation}

Perturbing \(y\) changes the ODE's initial state, and the velocity
field propagates this difference over the full integration
interval. When comparing \(y\) and \(y'=y+\delta\), we use the same
\(\epsilon\) for both restorations to measure observation sensitivity
with the source noise held fixed.

\subsubsection{Observation-Conditioned Posterior Sampling}
\label{sec:conditional_flow}

The second formulation samples the initial state from a standard Gaussian,
independently of the clean image and degraded observation, and supplies the
observation to the velocity field: $Z_0^{C}=\epsilon,\ \epsilon\sim\mathcal{N}(0,I).$
Using the path
\begin{equation}
    Z_t^{C}=(1-t)\epsilon+tX,
    \qquad
    U_t^{C}=X-\epsilon,
    \label{eq:conditional_path}
\end{equation}
we train an observation-conditioned velocity field
\(v_\theta(z,t;y)\) through
\begin{equation}
    \mathcal{L}_{\mathrm{CFM}}(\theta)
    =
    \mathbb{E}
    \left[
        \left\|
            v_\theta(Z_t^{C},t;Y)-U_t^{C}
        \right\|_2^2
    \right].
    \label{eq:conditional_fm_loss}
\end{equation}

The corresponding population-optimal field is
\begin{equation}
    v^*(z,t;y)
    =
    \mathbb{E}
    \left[
        X-\epsilon
        \mid Z_t^{C}=z,Y=y
    \right].
\end{equation}

Under the usual flow-matching assumptions and with an exact velocity field, integrating
\begin{equation}
    \frac{\mathrm{d}z_t}{\mathrm{d}t}
    =
    v_\theta(z_t,t;y),
    \qquad z_0=\epsilon,
    \label{eq:conditional_flow_ode}
\end{equation}
transports the base distribution toward the conditional clean-image
distribution \(p_{X\mid Y=y}\). The resulting restoration map is
\begin{equation}
    \mathcal{R}_{\mathrm{C}}(y;\epsilon)
    =
    \Phi^\theta_{0\rightarrow1,y}(\epsilon),
    \label{eq:conditional_restoration}
\end{equation}
where the subscript \(y\) emphasizes that the entire flow depends on the
observation.
Unlike Flow-from-$Y$, changing \(y\) does not alter the initial state when
the same \(\epsilon\) is used. Instead, it changes the predicted velocity at
every integration time. Observation perturbations may therefore be injected
continuously throughout the trajectory.

\subsection{Robustness Objective}
\label{sec:robustness_objective}

We study the input-to-reconstruction sensitivity of a stochastic restoration model
under fixed randomness. For a fixed randomness realization \(\xi\), observation \(y\), and
nonzero perturbation \(\delta\), define the finite-radius restoration gain
\begin{equation}
    G_{\mathcal{R}}(y,\delta;\xi)
    =
    \frac{
        \left\|
            \mathcal{R}(y+\delta;\xi)
            -
            \mathcal{R}(y;\xi)
        \right\|_2
    }{
        \left\|\delta\right\|_2
    }.
    \label{eq:end_to_end_gain}
\end{equation}
The corresponding worst-case gain within a radius \(\rho>0\) is
\begin{equation}
    G_{\mathcal{R}}^{(\rho)}(y;\xi)
    =
    \sup_{0<\|\delta\|_2\leq\rho}
    G_{\mathcal{R}}(y,\delta;\xi).
    \label{eq:worst_case_restoration_gain}
\end{equation}

\begin{figure}[htbp]
    \centering
    \includegraphics[width=\linewidth]
        {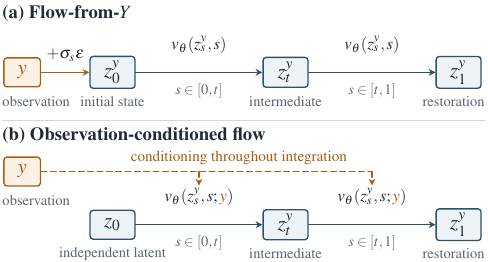}
    \caption{
        \textbf{Flow-based restoration formulations.}
        (a) Flow-from-$Y$ initializes the ODE with the
        noise-perturbed observation.
        (b) Observation-conditioned flow starts from a latent
        $z_0$ independent of $y$ and conditions the velocity
        field on $y$ throughout integration.
        Solid arrows between states denote ODE integration, and
        dashed arrows indicate observation conditioning.
    }
    \label{fig:flow_formulations}
\end{figure}

This quantity measures how strongly the complete restoration procedure can
amplify a bounded input change. It is distinct from sample diversity, since
the same random seed and numerical-solver configuration are used before and
after perturbation.
The two formulations induce different sensitivity pathways (see Figure~\ref{fig:flow_formulations}).
Accordingly, Flow-from-$Y$ requires control of the velocity field with
respect to its evolving state, whereas an observation-conditioned flow
requires control with respect to both its evolving state and the observation
condition. 
In the latter case, condition-wise sensitivity determines how
perturbations of the observation enter the dynamics, while state-wise
sensitivity governs how the resulting trajectory discrepancies propagate
during integration.
In Sec.~\ref{sec:stability_analysis}, we formalize these mechanisms and derive
bounds relating the Lipschitz behavior of the velocity field to the sensitivity
of the terminal restoration. These results subsequently motivate the
architecture and training objective introduced in Sec.~\ref{sec:lipsflower}.
\section{Stability Analysis}
\label{sec:stability_analysis}

We now characterize how perturbations of the degraded observation propagate
through the two formulations introduced in
Sec.~\ref{sec:observation_dependent_flows}. Our analysis considers the exact
continuous-time flow and uses the Euclidean norm throughout. We assume that
the vector fields are continuous in time and that the corresponding ODE
solutions exist on \(t\in[0,1]\)~\cite{chen2018neural}. All comparisons use the same source-noise
realization $\epsilon$, thereby isolating sensitivity to the observation from ordinary
sampling variability.

\subsection{Stability of Flow-from-$Y$ Restoration}
\label{sec:flow_y_stability}

In Flow-from-$Y$, the observation affects the restoration through the initial
state $z_0^y = y+\sigma_s\epsilon,$
while the velocity field itself is not explicitly conditioned on \(y\). 

\begin{proposition}[Stability of Flow-from-$Y$]
\label{prop:flow_y_stability}
Suppose that \(v_\theta\) is \(\rho\)-Lipschitz with respect to its state,
uniformly over time:
\begin{equation}
    \left\|
        v_\theta(a,t)-v_\theta(b,t)
    \right\|_2
    \leq
    \rho\left\|a-b\right\|_2
    \label{eq:state_lipschitz_assumption}
\end{equation}
for all \(a,b\in\mathbb{R}^{d}\) and \(t\in[0,1]\). Then, for any observations
\(y_1,y_2\) restored using the same \(\epsilon\),
\begin{equation}
    \left\|
        \mathcal{R}_{\mathrm{F}Y}(y_2;\epsilon)
        -
        \mathcal{R}_{\mathrm{F}Y}(y_1;\epsilon)
    \right\|_2
    \leq
    e^{\rho}
    \left\|y_2-y_1\right\|_2.
    \label{eq:flow_y_stability_bound}
\end{equation}
\end{proposition}

The proof is provided in the supplementary material. The bound in~\eqref{eq:flow_y_stability_bound}
is worst-case sharp under the stated Lipschitz assumption.
However, it can be conservative for a particular velocity field,
whose actual perturbation amplification may be much smaller
than \(e^\rho\).\footnote{For example, \(v_\theta(z,t)=\rho z\)
attains the bound, whereas \(v_\theta(z,t)=-\rho z\) has the same
Lipschitz constant but contracts perturbations by \(e^{-\rho}\).}

\subsection{Stability of Observation-Conditioned Flows}
\label{sec:conditional_stability}

In a conditional posterior sampler, the initial noise is independent of the
observation, but the observation affects the vector field throughout the
trajectory.
Consequently, observation sensitivity is governed by the dependence of the
field on both its evolving state and its condition.

\begin{proposition}[Stability of an observation-conditioned flow]
\label{prop:conditional_stability}
Suppose that \(v_\theta\) is uniformly \(\rho\)-Lipschitz in its state and uniformly
\(\rho_c\)-Lipschitz in its observation condition. That is, for every
\(t\in[0,1]\),
\begin{align}
    \left\|
        v_\theta(a,t;y)-v_\theta(b,t;y)
    \right\|_2
    &\leq
    \rho\left\|a-b\right\|_2,
    \label{eq:conditional_state_lipschitz}
    \\
    \left\|
        v_\theta(a,t;y_2)-v_\theta(a,t;y_1)
    \right\|_2
    &\leq
    \rho_c\left\|y_2-y_1\right\|_2.
    \label{eq:condition_lipschitz}
\end{align}
If \(\rho>0\), then, for any \(y_1,y_2\) and the same initial noise
\(\epsilon\),
\begin{equation}
    \left\|
        \mathcal{R}_{\mathrm{C}}(y_2;\epsilon)
        -
        \mathcal{R}_{\mathrm{C}}(y_1;\epsilon)
    \right\|_2
    \leq
    \frac{\rho_c}{\rho}
    \left(e^{\rho}-1\right)
    \left\|y_2-y_1\right\|_2.
    \label{eq:conditional_stability_bound}
\end{equation}
\end{proposition}

We give the proof in the supplementary material. The bound in~\eqref{eq:conditional_stability_bound}
is also worst-case sharp under the stated assumptions, although
it may be conservative for a particular velocity field.
\footnote{Equality is attained for the scalar
field \(v_\theta(z,t;y)=\rho z+\rho_c y\) when both trajectories
start from the same initial noise.}
In this bound, the factor \(\rho_c\) controls how
strongly an observation perturbation is injected into the velocity field,
whereas \(\rho\) controls how the induced trajectory discrepancy is
subsequently propagated. Controlling only \(\rho_c\) is therefore insufficient:
even a small condition-induced discrepancy can be strongly amplified by
state-sensitive dynamics. 

\subsection{Implications for Model Design}
\label{sec:stability_design_implications}

The stability analysis yields three design requirements. \emph{First}, both restoration
formulations require control of the velocity field with respect to the evolving
state. \emph{Second}, an observation-conditioned flow must additionally control its
dependence on the conditioning image. \emph{Third}, control should apply throughout
the transport interval, because sensitivity accumulated at intermediate states
affects the terminal reconstruction.
The next section explains how LipsFlower addresses these three
requirements through architectural controls and gain regularization.

\section{LipsFlower}
\label{sec:lipsflower}

Guided by our stability analysis, LipsFlower combines a
Lipschitz-aware velocity network with finite-radius gain
regularization. We instantiate this framework for both
Flow-from-$Y$ and observation-conditioned posterior sampling,
targeting state sensitivity in both formulations and
observation sensitivity in the conditional formulation.

\subsection{Lipschitz-Aware Velocity Network}
\label{sec:liphdit}

Our velocity network, \emph{LipHDiT}, keeps the overall
structure of HDiT~\cite{crowson2024scalable} and adds controls
to limit amplification. Linear projections use approximate
spectral normalization~\cite{miyato2018spectral} with a safety
margin and full-precision power iteration. We include safeguards
against degenerate iteration vectors, including for zero-initialized
weights. We add a denominator floor to RMSNorm~\cite{zhang2019root}
and bound the time-dependent modulation used for adaptive
conditioning~\cite{crowson2024scalable,peebles2023scalable}.
Our query-key normalization also uses a denominator floor,
building on normalized-attention
designs~\cite{henry2020query,crowson2024scalable}.
To control amplification in the residual branches and decoder skip connections, we use convex
aggregation~\cite{cisse2017parseval}.
For conditional sampling, separate
state and observation projection stems feed a controlled fusion
module before the shared backbone. These controls aim to reduce
amplification without requiring every component to be nonexpansive. More architectural details are given in
\cref{app:lipsflower_details} of the supplementary material.

\subsection{Finite-Radius Gain Regularization}
\label{sec:gain_regularization}

\begin{figure*}[htbp]
    \centering
    \includegraphics[width=\linewidth]
        {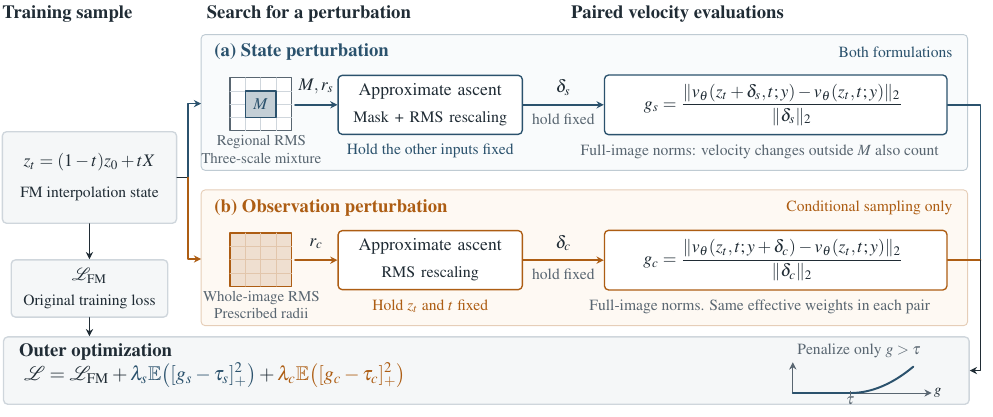}
    \caption{\textbf{Finite-radius gain regularization.}
    Adversarial searches perturb local state regions and,
    for conditional sampling (marked in orange), the whole observation.
    Squared hinge penalties on velocity gains above prescribed
    thresholds are added to the flow-matching loss.}
    \label{fig:gain_regularization}
\end{figure*}

We also adapt adversarial Lipschitz
regularization~\cite{terjek2019adversarial} to restoration
velocity fields. It searches for perturbations that cause large
velocity changes and penalizes gains above a chosen threshold.
We apply it at sampled flow-matching interpolation
states~\cite{lipman2023flow}, so regularization does not require
full ODE rollouts. For observation-conditioned fields, our
stability analysis motivates controlling sensitivity to both
the evolving state and the observation. We measure the two
finite-radius gains separately:
\begin{equation}
\begin{aligned}
    g_s &=
    \frac{
        \|v_\theta(z_t+\delta_s,t;y)
          -v_\theta(z_t,t;y)\|_2
    }{\|\delta_s\|_2},
    \\
    g_c &=
    \frac{
        \|v_\theta(z_t,t;y+\delta_c)
          -v_\theta(z_t,t;y)\|_2
    }{\|\delta_c\|_2}.
\end{aligned}
\label{eq:velocity_gains}
\end{equation}

For Flow-from-$Y$, we use only \(g_s\), with no explicit
observation condition.
We use localized state perturbations to discourage amplification of spatially concentrated disturbances. However, the gains in Eq.~\eqref{eq:velocity_gains} use full-image norms,
so velocity changes outside the perturbed region also count. The root-mean-square (RMS) magnitude of the perturbations within the region
is drawn from a mixture of small, moderate, and severe scales,
with more weight on small perturbations. Starting from random
noise, we use projected gradient ascent~\cite{madry2018towards}
to approximately maximize the velocity discrepancy.
After each update, we restrict the perturbation to its sampled
region and rescale it to the chosen RMS magnitude.
For conditional sampling, we search separately for observation
perturbations at prescribed whole-image RMS radii. This addresses distributed measurement uncertainty.
With these adversarial perturbations, we optimize
\begin{equation}
\begin{aligned}
    \mathcal{L}
    ={}& \mathcal{L}_{\mathrm{FM}}
    + \lambda_s
      \mathbb{E}\!\left([g_s-\tau_s]_+^2\right)
    + \lambda_c
      \mathbb{E}\!\left([g_c-\tau_c]_+^2\right),
\end{aligned}
\label{eq:lipsflower_objective}
\end{equation}
where \([a]_+=\max(a,0)\), \(\tau_s\) and \(\tau_c\) are gain
thresholds, and \(\lambda_s\) and \(\lambda_c\) weight the
sensitivity penalties relative to velocity regression.
The expectations are over sampled training states and
perturbation searches, with Flow-from-$Y$ using only the state penalty. 

These architectural controls and sampled gain penalties target the sensitivities identified by our analysis. As illustrated in \cref{fig:teaser}, small adversarial perturbations
to the observation can lead to noticeable reconstruction changes.
LipsFlower reduces these changes for both Flow-from-$Y$ and observation-conditioned posterior sampling. The supplementary material
describes the perturbation distributions, search settings, and
regularization schedule (see \cref{app:lipsflower_details}).
\section{Experiments}
\label{sec:experiments}

\begin{table*}[t]
    \centering
    \caption{Adversarial robustness under SPSA attacks on CelebA-Test.
     LPIPS drift is reported in units of $10^{-3}$.
     $\mathrm{CVaR}_{95}\,G$ is the mean of the worst 5\% among all test images. The blocks (I) and (II) present the comparison sets for the Flow-from-$Y$ and observation-conditioned formulations of LipsFlower, respectively.
     Within each category, \textbf{bold} and \underline{underlined} values indicate the best and second-best results, respectively.}
    \label{tab:spsa_robustness}
    \begingroup
    \footnotesize
    \setlength{\tabcolsep}{1.6pt}
    \renewcommand{\arraystretch}{1.12}
    \begin{tabular}{clrrrrrrrrrr}
        \toprule
         \multicolumn{2}{c}{\multirow{2}{*}{Method}} & \multicolumn{5}{c}{SPSA--$\ell_2$ (RMS budget)}
        & \multicolumn{5}{c}{SPSA--$\ell_\infty$} \\
        \cmidrule(lr){3-7}\cmidrule(lr){8-12}
        & & Mean $G$ $\downarrow$ & P95 $G$ $\downarrow$ & $\mathrm{CVaR}_{95}\,G$ $\downarrow$ & Max $G$ $\downarrow$ & LPIPS drift $\downarrow$
        & Mean $G$ $\downarrow$ & P95 $G$ $\downarrow$ & $\mathrm{CVaR}_{95}\,G$ $\downarrow$ & Max $G$ $\downarrow$ & LPIPS drift $\downarrow$ \\
        \midrule
        \multirow{8}{*}{(I)} &
        CodeFormer                 & 2.44 & 4.04 & 4.577 & 7.17  & 63.375  & 2.91 & 4.82 & 5.495 & 9.35  & 53.101 \\
        & GFPGAN                   & 3.01 & 5.13 & 5.664 & 7.53  & 110.573 & 3.60 & 6.45 & 7.122 & 9.57  & 98.214 \\
        & RestoreFormer++          & 1.77 & 3.75 & 4.445 & 6.79  & 44.146  & 2.10 & 4.56 & 5.406 & 8.53  & 38.113 \\
        & OSDFace                  & 4.99 & 6.49 & 6.898 & 8.13  & 186.731 & 5.83 & 7.77 & 8.262 & 9.87  & 155.168 \\
        & PMRF                     & \textbf{0.47} & 1.11 & 1.455 & 2.91  & \textbf{9.033} & \textbf{0.47} & 1.03 & 1.355 & 2.96  & \textbf{5.454} \\
        & ELIR                     & 12.82 & 23.03 & 25.700 & 36.00 & 459.179 & 12.99 & 22.98 & 25.643 & 35.37 & 451.283 \\
        & Flow-from-$Y$            & 0.52 & \underline{0.74} & \underline{0.849} & \underline{1.74} & 15.803 & 0.52 & \underline{0.74} & \underline{0.830} & \underline{1.66} & 9.188 \\ \rowcolor{gray!18} \cellcolor{white}
        &  LipsFlower (w/ Flow-from-$Y$) & \textbf{0.47} & \textbf{0.67} & \textbf{0.747} & \textbf{1.16} & \underline{11.649} & \underline{0.48} & \textbf{0.66} & \textbf{0.727} & \textbf{1.12} & \underline{7.360} \\
        \midrule
        \midrule
         &
        DiffBIR                    & 0.90 & 2.05 & 2.892 & 10.28 & 14.711 & 0.96 & 2.22 & 3.296 & 10.67 & 10.459 \\
        & FlowIE                   & 0.64 & 1.45 & 1.856 & 3.51 & 7.635 & 0.65 & 1.39 & 1.834 & 3.75 & 4.935 \\
        & BayesFlow                & \textbf{0.23} & \textbf{0.35} & \underline{0.412} & \underline{0.95} & \underline{2.901}
                                   & \textbf{0.25} & \textbf{0.36} & \underline{0.433} & \underline{0.73} & \underline{2.024} \\
        \rowcolor{gray!18} \cellcolor{white} \multirow{-4}{*}{(II)} & LipsFlower (w/ BayesFlow)   & \underline{0.26} & \underline{0.36} & \textbf{0.406} & \textbf{0.56} & \textbf{1.648}
                                   & \underline{0.28} & \underline{0.38} & \textbf{0.415} & \textbf{0.56} & \textbf{1.088} \\
        \bottomrule
    \end{tabular}
    \endgroup
\end{table*}

We evaluate LipsFlower on blind face restoration, testing
whether it reduces the amplification of observation
perturbations compared with its flow-matching baselines
and how this affects restoration quality. The supplementary material reports an ablation study of architectural controls and gain regularization, individually and jointly, in~\cref{app:component_ablation}.

\subsection{Experimental Setup}
\label{sec:experimental_setup}

\paragraph{Data and comparisons.}
Our models are trained on FFHQ~\cite{karras2019style} at
$512\times512$ resolution. Evaluation uses a shared set
of $3{,}000$ degraded--clean image pairs from
CelebA-Test~\cite{wang2021towards} at the same resolution.
We compare each LipsFlower variant with its corresponding
baseline: Flow-from-$Y$~\cite{ohayon2025posterior} and an
observation-conditioned sampler implemented using
BayesFlow~\cite{kuhmichel2026bayesflow}.
BayesFlow is a framework for amortized Bayesian inference.
Our adaptation uses its conditional flow-matching component
to learn a velocity field that transports Gaussian noise
toward clean images, with the degraded observation conditioned
throughout integration.

The baselines are grouped by their inference pathway.
Group~(I) includes direct restoration methods
(CodeFormer~\cite{zhou2022towards},
GFPGAN~\cite{wang2021towards}, and
RestoreFormer++~\cite{wang2023restoreformer++}) and methods
that refine or transport an image or latent representation
derived from the observation
(OSDFace~\cite{wang2025osdface},
PMRF~\cite{ohayon2025posterior},
ELIR~\cite{cohen2025efficient}, and Flow-from-$Y$).
Group~(II) includes methods that start from
observation-independent noise and condition their sampling
dynamics on the observation
(DiffBIR~\cite{lin2024diffbir},
FlowIE~\cite{zhu2024flowie}, and our BayesFlow adaptation).
Each LipsFlower variant appears alongside its corresponding
flow-matching baseline.
Training and method-specific inference settings are given
in the supplementary material (\cref{app:training_inference}).

\paragraph{Restoration metrics.}
We measure reconstruction fidelity using RGB PSNR~\cite{hore2010image} and perceptual similarity using VGG-based
LPIPS~\cite{zhang2018unreasonable}. We report
FID~\cite{heusel2017gans} and generative precision~\cite{kynkaanniemi2019improved}
against a common FFHQ reference distribution.
Face-specific fidelity is assessed using the angular
distance between ArcFace embeddings, denoted DEG~\cite{deng2019arcface}
and measured in degrees, and the mean distance
between predicted facial landmarks, denoted LMD~\cite{wang2019adaptive}
and measured in image pixels. All methods use the
same evaluation images and metric implementations.

\paragraph{Adversarial evaluation.}
We evaluate observation sensitivity using simultaneous
perturbation stochastic approximation
(SPSA)~\cite{spall1992multivariate, uesato2018adversarial}, which requires
only queries to the restoration procedure. For observations (height \(H\) and width \(W\)) with pixel values scaled to $[0,1]$, we consider nominal budgets $\|\delta\|_2/\sqrt{3HW}\leq0.01$ and
$\|\delta\|_\infty\leq0.01$.
Searches use $20$ iterations with two queries each, probe radius $0.0025$, and step size $0.002$. For each method $\mathcal{R}$, amplification is measured using $G=G_{\mathcal{R}}(y,\delta;\xi)$, defined in~\eqref{eq:end_to_end_gain}. Randomness $\xi$ is fixed across paired evaluations.
For both attacks, we report the mean, 95th percentile, maximum, and \(\mathrm{CVaR}_{95}\) of the gains across all test images. \(\mathrm{CVaR}_{95}\) denotes the Conditional Value at Risk at the 95th percentile, computed as the mean of the largest \(5\%\) of gains. We also report LPIPS drift between clean and attacked reconstructions.

\subsection{Robustness to Observation Perturbations}
\label{sec:restoration_robustness}

We first examine the primary objective of LipsFlower, which is to make restorations less sensitive to small changes in the degraded observation.
Table~\ref{tab:spsa_robustness} measures how much an
input perturbation changes the reconstructed image.
Lower gain indicates less amplification, while lower
LPIPS drift means that the perturbed reconstruction
remains perceptually closer to the original output.

\paragraph{Flow-from-$Y$.}
LipsFlower improves every reported robustness measure
relative to the Flow-from-$Y$ baseline under both
attack constraints. Under the RMS-constrained attack,
mean gain decreases from $0.52$ to $0.47$.
The improvement also extends to the most sensitive images. The average gain among the worst $5\%$
of cases, measured by $\mathrm{CVaR}_{95}$,
decreases from $0.849$ to $0.747$, while the maximum
observed gain decreases from $1.74$ to $1.16$.
These correspond to reductions of approximately
$12\%$ and $33\%$, respectively.
The $\ell_\infty$ attack produces a similar pattern.
The average gain among the worst $5\%$ of cases
decreases from $0.830$ to $0.727$, and the maximum
decreases from $1.66$ to $1.12$.
These reductions are accompanied by lower perceptual
drift: LPIPS drift decreases by approximately $26\%$
under the RMS-constrained attack and $20\%$ under
the $\ell_\infty$ attack.
Thus, the improvement concerns both numerical
amplification and perceptual changes in the output.
The literature comparisons provide further context.
PMRF achieves a similar mean gain and lower LPIPS
drift, whereas LipsFlower achieves lower
95th-percentile, tail-average, and maximum gains.

\paragraph{Observation-conditioned sampling.}
For conditional sampling, LipsFlower primarily reduces
large observed changes and perceptual drift.
Relative to BayesFlow, maximum gain decreases from
$0.95$ to $0.56$ under the RMS-constrained attack
and from $0.73$ to $0.56$ under the $\ell_\infty$
attack. LPIPS drift decreases by approximately
$43\%$ and $46\%$, respectively.
However, mean gain and the 95th percentile increase
slightly. The conditional formulation therefore
improves some aspects of robustness without uniformly
reducing sensitivity across the test set.

\subsection{Restoration Quality}
\label{sec:restoration_quality}

With Flow-from-$Y$, LipsFlower reduces perturbation amplification while PSNR on unperturbed observations drops by only $0.31$\,dB relative to the baseline (Table~\ref{tab:perceptual_quality} in \cref{app:restoration_quality}). FID rises from $37.05$ to $40.77$, close to that of RestoreFormer++ ($40.61$). With BayesFlow, the reduction in maximum observed amplification and perceptual drift comes at a larger cost to restoration quality. PSNR drops by $2.32$\,dB and FID rises from $40.24$ to $57.04$. We provide the full comparison in the supplementary material (\cref{app:restoration_quality}).

Taken together, the two formulations demonstrate different robustness--quality trade-offs. \emph{Flowing directly from the observation offers a better balance between robustness and restoration quality, whereas conditioning on the observation reduces maximum amplification and perceptual drift at a moderately higher cost to perceptual fidelity.}

\section{Limitations}
\label{sec:limitations}

Our results show that Lipschitz-aware flow matching can reduce perturbation amplification, although the effect depends on the restoration formulation. In particular, LipsFlower (w/ Flow-from-$Y$) gives a better balance between robustness and reconstruction quality,
whereas observation-conditioned sampling loses more perceptual quality and shows less consistent robustness gains.
At the same time, the architectural controls and finite-radius regularization do not provide a global Lipschitz guarantee.
Therefore, the stability results still depend on the assumptions used in the analysis. In addition, our experiments are limited to blind face restoration.
Evaluating other inverse problems and stronger perturbations are good directions for future work.
\section{Conclusion}
\label{sec:conclusion}

We introduced LipsFlower to make flow-based image restoration less sensitive to input perturbations. 
Our method combines Lipschitz-aware architectural controls with finite-radius gain regularization, while our stability analysis explains how observation perturbations enter and propagate through two flow-matching formulations. 
On blind face restoration, LipsFlower reduces perturbation amplification, with Flow-from-$Y$ giving the better balance between robustness and reconstruction quality. 
Overall, the results show that controlling sensitivity in both the architecture and training objective can improve the stability of generative image restoration.

\section{Acknowledgment}
All experiments were performed on the compute cluster of the \emph{Lamarr Institute for Machine Learning and Artificial Intelligence}.

{
    \small
    \bibliographystyle{ieeenat_fullname}
    \bibliography{main}
}

\clearpage
\setcounter{page}{1}
\maketitlesupplementary


\appendix
\etocdepthtag.toc{appendix}

\begingroup
    \renewcommand{\contentsname}{Contents}
    \setcounter{tocdepth}{2} 

    \etocsettagdepth{main}{none}           
    \etocsettagdepth{appendix}{subsection} 

    \tableofcontents
\endgroup

\section{Proofs of the Stability Results}
\label{app:stability_proofs}

This section provides the complete proofs of the stability results stated in
the main paper. For completeness, we first recall the
Gr\"onwall--Bellman inequality~\cite{gronwall1919note,bellman1943stability,PACHPATTE1973758} used in both derivations and then restate each
proposition before presenting its proof. All norms are Euclidean, and all
trajectory comparisons use the same source-noise realization unless stated
otherwise.

\subsection{Gr\"onwall--Bellman Inequality}
\label{app:gronwall}

The following standard integral inequality controls the growth of a
nonnegative function and is the principal tool used to translate local
Lipschitz control of the velocity field into an end-to-end stability bound.

\begin{lemma}[Gronwall--Bellman inequality]
\label{lem:gronwall}
Let $d(t)$ and $h(t)$ be nonnegative, continuous functions on $\mathcal{I}=[a,b]$.
Assume that $d_0$ is a nonnegative constant and that
$$
d(t) \le d_0 + \int_a^t h(s)d(s)\ \mathrm{d}s,\quad \text{for all} \quad t \in \mathcal{I}.
$$
Then
$$
d(t) \le d_0\exp\left(\int_a^t h(s)\ \mathrm{d}s\right),\quad \text{for all} \quad t \in \mathcal{I}.
$$
\end{lemma}


\subsection{Proof of Proposition \ref{prop:flow_y_stability}}

We first consider Flow-from-$Y$, where the degraded observation determines
the initial ODE state. We restate the proposition to make the supplementary
material self-contained.

\begin{restatedproposition1}[Stability of Flow-from-$Y$]
Suppose that \(v_\theta\) is \(\rho\)-Lipschitz with respect to its state,
uniformly over time:
\begin{equation}
    \left\|
        v_\theta(a,t)-v_\theta(b,t)
    \right\|_2
    \leq
    \rho\left\|a-b\right\|_2
\end{equation}
for all \(a,b\in\mathbb{R}^{d}\) and \(t\in[0,1]\). Then, for any observations
\(y_1,y_2\) restored using the same \(\epsilon\),
\begin{equation}
    \left\|
        \mathcal{R}_{\mathrm{F}Y}(y_2;\epsilon)
        -
        \mathcal{R}_{\mathrm{F}Y}(y_1;\epsilon)
    \right\|_2
    \leq
    e^{\rho}
    \left\|y_2 -y_1 \right\|_2.
\end{equation}
\end{restatedproposition1}

\begin{proof}
Let \(z^1:[0,1]\rightarrow\mathbb{R}^{d}\) and
\(z^2:[0,1]\rightarrow\mathbb{R}^{d}\) denote the Flow-from-$Y$
trajectories associated with observations \(y_1\) and \(y_2\), respectively,
using the same source-noise realization \(\epsilon\). They satisfy
\begin{align}
    \frac{\mathrm{d}z_t^1}{\mathrm{d}t}
    &=
    v_\theta(z_t^1,t),
    &
    z_0^1
    &=
    y_1 +\sigma_s\epsilon,
    \label{eq:proof_flow_y_trajectory_1}
    \\
    \frac{\mathrm{d}z_t^2}{\mathrm{d}t}
    &=
    v_\theta(z_t^2,t),
    &
    z_0^2
    &=
    y_2 +\sigma_s\epsilon,
    \label{eq:proof_flow_y_trajectory_2}
\end{align}
and their terminal states are
\begin{equation}
    z_1^i
    =
    \mathcal{R}_{\mathrm{F}Y}(y_i;\epsilon),\quad \text{for}\quad i \in \lbrace 1, 2 \rbrace.
    \label{eq:proof_flow_y_terminal_states}
\end{equation}
Integrating Eqs.~\eqref{eq:proof_flow_y_trajectory_1} and
\eqref{eq:proof_flow_y_trajectory_2} from \(0\) to \(t\) gives
\begin{align}
    z_t^1
    &=
    z_0^1+\int_0^t v_\theta(z_s^1,s)\,\mathrm{d}s,
    \\
    z_t^2
    &=
    z_0^2+\int_0^t v_\theta(z_s^2,s)\,\mathrm{d}s.
\end{align}
Subtracting the first equality from the second yields
\begin{equation}
    z_t^2-z_t^1
    =
    z_0^2-z_0^1
    +
    \int_0^t
    \left[
        v_\theta(z_s^2,s)-v_\theta(z_s^1,s)
    \right]\mathrm{d}s.
    \label{eq:proof_flow_y_difference}
\end{equation}
Taking the Euclidean norm and applying the triangle inequality,
\begin{align}
    \left\|z_t^2-z_t^1\right\|_2
    \leq\;&
    \left\|z_0^2-z_0^1\right\|_2
    \nonumber\\
    &+
    \int_0^t
    \left\|
        v_\theta(z_s^2,s)-v_\theta(z_s^1,s)
    \right\|_2
    \mathrm{d}s.
    \label{eq:proof_flow_y_triangle}
\end{align}
By the assumed uniform \(\rho\)-Lipschitz continuity of \(v_\theta\) with
respect to its state,
\begin{equation}
    \left\|
        v_\theta(z_s^2,s)-v_\theta(z_s^1,s)
    \right\|_2
    \leq
    \rho\left\|z_s^2-z_s^1\right\|_2.
    \label{eq:proof_flow_y_apply_lipschitz}
\end{equation}
Substituting Eq.~\eqref{eq:proof_flow_y_apply_lipschitz} into
Eq.~\eqref{eq:proof_flow_y_triangle} gives
\begin{equation}
    \left\|z_t^2-z_t^1\right\|_2
    \leq
    \left\|z_0^2-z_0^1\right\|_2
    +
    \rho
    \int_0^t
    \left\|z_s^2-z_s^1\right\|_2
    \mathrm{d}s.
    \label{eq:proof_flow_y_gronwall_form}
\end{equation}
Define
\begin{equation}
    d(t)
    \coloneqq
    \left\|z_t^2-z_t^1\right\|_2.
\end{equation}
The function \(d\) is nonnegative and continuous, and
Eq.~\eqref{eq:proof_flow_y_gronwall_form} becomes
\begin{equation}
    d(t)
    \leq
    d(0)+\int_0^t \rho d(s)\,\mathrm{d}s.
    \label{eq:proof_flow_y_scalar_inequality}
\end{equation}
Applying the Gr\"onwall--Bellman inequality (~\cref{app:gronwall}) with
\(d_0=d(0)\) and \(h(s)=\rho\) yields
\begin{equation}
    d(t)
    \leq
    d(0)
    \exp\left(\int_0^t\rho\,\mathrm{d}s\right)
    =
    e^{\rho t}d(0).
    \label{eq:proof_flow_y_gronwall_result}
\end{equation}

Because the two trajectories use the same source-noise realization,
\begin{equation}
    z_0^2-z_0^1
    =
    \left(y_2 +\sigma_s\epsilon\right)
    -
    \left(y_1 +\sigma_s\epsilon\right)
   =
    y_2 -y_1 .
    \label{eq:proof_flow_y_noise_cancellation}
\end{equation}
Therefore,
\begin{equation}
    d(0)
    =
    \left\|y_2 -y_1 \right\|_2.
    \label{eq:proof_flow_y_initial_distance}
\end{equation}
Evaluating Eq.~\eqref{eq:proof_flow_y_gronwall_result} at \(t=1\), and using
Eq.~\eqref{eq:proof_flow_y_terminal_states}, gives
\begin{align}
    &
    \left\|
        \mathcal{R}_{\mathrm{F}Y}(y_2;\epsilon)
        -
        \mathcal{R}_{\mathrm{F}Y}(y_1;\epsilon)
    \right\|_2
    \nonumber\\
    &\qquad=
    \left\|z_1^2-z_1^1\right\|_2
    \leq
    e^{\rho}
    \left\|y_2 -y_1 \right\|_2,
\end{align}
which proves Proposition~\ref{prop:flow_y_stability}.
\end{proof}

The preceding result controls perturbations that enter through the initial
state. We next consider the setting in which the initial state is fixed under common-noise coupling, while the observation affects the velocity field throughout the transport trajectory.

\subsection{Proof of Proposition~\ref{prop:conditional_stability}}

For an observation-conditioned flow, robustness depends on the sensitivity of
the velocity field with respect to both its evolving state and its condition.
We again restate the proposition before giving the complete derivation.

\begin{restatedproposition2}[Stability of an observation-conditioned flow]
Suppose that \(v_\theta\) is uniformly \(\rho\)-Lipschitz in its state and uniformly \(\rho_c\)-Lipschitz in its observation condition. That is, for every
\(t\in[0,1]\),
\begin{align}
    \left\|
        v_\theta(a,t;y)-v_\theta(b,t;y)
    \right\|_2
    &\leq
    \rho\left\|a-b\right\|_2,
    \\
    \left\|
        v_\theta(a,t;y_2)-v_\theta(a,t;y_1)
    \right\|_2
    &\leq
    \rho_c\left\|y_2-y_1\right\|_2.
\end{align}
If \(\rho>0\), then, for any \(y_1,y_2\) and the same initial noise
\(\epsilon\),
\begin{equation}
    \left\|
        \mathcal{R}_{\mathrm{C}}(y_2;\epsilon)
        -
        \mathcal{R}_{\mathrm{C}}(y_1;\epsilon)
    \right\|_2
    \leq
    \frac{\rho_c}{\rho}
    \left(e^{\rho}-1\right)
    \left\|y_2-y_1\right\|_2.
\end{equation}
\end{restatedproposition2}


\begin{proof}
Let \(z^1:[0,1]\rightarrow\mathbb{R}^{d}\) and
\(z^2:[0,1]\rightarrow\mathbb{R}^{d}\) denote two conditional flow
trajectories associated with observations \(y_1\) and \(y_2\), respectively.
Let \(\epsilon \sim\mathcal{N}(0,I)\) be the source-noise realization for both trajectories:
\begin{align}
    \frac{\mathrm{d}z_t^1}{\mathrm{d}t}
    &=
    v_\theta(z_t^1,t;y_1),
    &
    z_0^1
    &=
    \epsilon,
    \label{eq:proof_conditional_trajectory_1}
    \\
    \frac{\mathrm{d}z_t^2}{\mathrm{d}t}
    &=
    v_\theta(z_t^2,t;y_2),
    &
    z_0^2
    &=
    \epsilon.
    \label{eq:proof_conditional_trajectory_2}
\end{align}
Their terminal states satisfy
\begin{equation}
    z_1^i
    =
    \mathcal{R}_{\mathrm{C}}(y_i;\epsilon),
    \qquad \text{for}\quad i\in\{1,2\}.
    \label{eq:proof_conditional_terminal_states}
\end{equation}
Integrating Eqs.~\eqref{eq:proof_conditional_trajectory_1} and
\eqref{eq:proof_conditional_trajectory_2} from \(0\) to \(t\) gives
\begin{align}
    z_t^1
    &=
    \epsilon
    +
    \int_0^t
    v_\theta(z_s^1,s;y_1)\,\mathrm{d}s,
    \\
    z_t^2
    &=
    \epsilon
    +
    \int_0^t
    v_\theta(z_s^2,s;y_2)\,\mathrm{d}s.
\end{align}
Subtracting the first equality from the second yields
\begin{equation}
\begin{aligned}
    z_t^2-z_t^1 = \int_0^t
        v_\theta(z_s^2,s;y_2)
        -
        v_\theta(z_s^1,s;y_1)
    \,\mathrm{d}s.
\end{aligned}
\label{eq:proof_conditional_difference}
\end{equation}
Taking norms and applying the triangle inequality,
\begin{equation}
\begin{aligned}
    &\left\|z_t^2-z_t^1\right\|_2
    \leq\; \\&
    \int_0^t
    \left\|
        v_\theta(z_s^2,s;y_2)
        -
        v_\theta(z_s^1,s;y_1)
    \right\|_2
    \,\mathrm{d}s.
\end{aligned}
\label{eq:proof_conditional_triangle}
\end{equation}
To separate the effects of the state and the observation, we add and subtract
\(v_\theta(z_s^1,s;y_2)\): 
\begin{equation}
\begin{aligned}
&
\left\|
    v_\theta(z_s^2,s;y_2)
    -
    v_\theta(z_s^1,s;y_1)
\right\|_2
\\
\quad=&
\left\|
    v_\theta(z_s^2,s;y_2)
    -
    v_\theta(z_s^1,s;y_2)
    +
    v_\theta(z_s^1,s;y_2)
    -
    v_\theta(z_s^1,s;y_1)
\right\|_2
\\
\quad\leq&
\left\|
    v_\theta(z_s^2,s;y_2)
    -
    v_\theta(z_s^1,s;y_2)
\right\|_2
\\
\qquad&+
\left\|
    v_\theta(z_s^1,s;y_2)
    -
    v_\theta(z_s^1,s;y_1)
\right\|_2.
\end{aligned}
\label{eq:proof_conditional_add_subtract}
\end{equation}
The first term changes the state while holding the condition fixed, whereas
the second changes the condition while holding the state fixed. By the
assumed Lipschitz continuity in these two arguments,
\begin{align}
    \left\|
        v_\theta(z_s^2,s;y_2)
        -
        v_\theta(z_s^1,s;y_2)
    \right\|_2
    &\leq
    \rho\left\|z_s^2-z_s^1\right\|_2,
    \label{eq:proof_conditional_state_bound}
    \\
    \left\|
        v_\theta(z_s^1,s;y_2)
        -
        v_\theta(z_s^1,s;y_1)
    \right\|_2
    &\leq
    \rho_c\left\|y_2-y_1\right\|_2.
    \label{eq:proof_conditional_condition_bound}
\end{align}
Combining Eqs.~\eqref{eq:proof_conditional_triangle}--%
\eqref{eq:proof_conditional_condition_bound} gives
\begin{equation}
\begin{aligned}
    \left\|z_t^2-z_t^1\right\|_2
    \leq\;&
    \rho
    \int_0^t
    \left\|z_s^2-z_s^1\right\|_2
    \,\mathrm{d}s
    \\
    +& \,
    \rho_c
    \int_0^t
    \left\|y_2-y_1\right\|_2
    \,\mathrm{d}s \\
    = & 
    \rho
    \int_0^t\left(
    \left\|z_s^2-z_s^1\right\|_2 + \frac{\rho_c}{\rho}\Delta y\right)\,
    \,\mathrm{d}s,
\end{aligned}
\label{eq:proof_conditional_combined_bound}
\end{equation}
with $\Delta y \coloneqq\left\|y_2-y_1\right\|_2$. Next, we define
\begin{equation}
    \begin{aligned}
        d(t) \coloneqq  \left\|z_t^2-z_t^1\right\|_2 + \frac{\rho_c}{\rho}\Delta y,
    \end{aligned}
\end{equation}
with $d(0) = \frac{\rho_c}{\rho}\Delta y$, since \(z_0^2 = z_0^1 = \epsilon\). Then, by Eq. \eqref{eq:proof_conditional_combined_bound}, we obtain
\begin{equation}
    \begin{aligned}
        d(t) \le d(0) + \int_0^t \rho
    d(s)\,
    \,\mathrm{d}s.
    \end{aligned}
\end{equation}
Then, we can apply the Gr\"onwall-Bellman inequality, to derive
\begin{equation}
    \begin{aligned}
    d(t) \le d(0) e^{ \rho t }.
    \end{aligned}
\end{equation}
Using the definition of $d(t)$ and $\Delta y$, we obtain
\begin{align*}
     \left\|z_t^2-z_t^1\right\|_2 \le \frac{\rho_c}{\rho}\left( e^{ \rho t } - 1\right) \left\|y_2-y_1\right\|_2.
\end{align*}
Finally, setting \(t=1\) and using
Eq.~\eqref{eq:proof_conditional_terminal_states} gives
\begin{equation}
\begin{aligned}
&
\left\|
    \mathcal{R}_{\mathrm{C}}(y_2;\epsilon)
    -
    \mathcal{R}_{\mathrm{C}}(y_1;\epsilon)
\right\|_2
\\
&\qquad=\left\|z_1^2-z_1^1\right\|_2\leq
\frac{\rho_c}{\rho}
\left(e^{\rho}-1\right)
\left\|y_2-y_1\right\|_2,
\end{aligned}
\end{equation}
which proves the proposition.
\end{proof}


\begin{table*}[t]
    \centering
    \caption{Clean reconstruction and perceptual quality on CelebA-Test at $512\times512$ resolution. The blocks (I) and (II) present the comparison sets for the Flow-from-$Y$ and observation-conditioned formulations of LipsFlower, respectively. Within each category, \textbf{bold} and \underline{underlined} values indicate the best and second-best results, respectively.}
    \label{tab:perceptual_quality}
    \begingroup
    \small
    \setlength{\tabcolsep}{3pt}
    \renewcommand{\arraystretch}{1.12}
    \begin{tabular}{clrrrrrr}
        \toprule
        \multicolumn{2}{c}{Method} & PSNR $\uparrow$ & LPIPS $\downarrow$
        & FID $\downarrow$ & Precision $\uparrow$
        & DEG $\downarrow$ & LMD $\downarrow$ \\
        \midrule
        \multirow{8}{*}{(I)} & CodeFormer & 25.146 & \textbf{0.343} & 52.521 & 0.3512 & 37.282 & 2.470 \\
        & GFPGAN & 24.986 & 0.364 & 46.160 & 0.3212 & 36.036 & 2.444 \\
        & RestoreFormer++ & 25.313 & \underline{0.344} & 40.611 & \underline{0.4089} & \textbf{29.634} & 2.043 \\
        & OSDFace & 23.381 & 0.364 & 43.567 & \textbf{0.4631} & 33.907 & 2.158 \\
        & PMRF & \textbf{25.821} & 0.367 & \textbf{35.842} & 0.3855 & 31.027 & \underline{2.037} \\
        & ELIR & \underline{25.817} & 0.374 & 39.944 & 0.2441 & 34.258 & 2.308 \\
        & Flow-from-$Y$ & 25.784 & 0.350 & \underline{37.050} & 0.3797 & \underline{30.602} & \textbf{1.995} \\
        \rowcolor{gray!18} \cellcolor{white} & LipsFlower (w/ Flow-from-$Y$) & 25.475 & 0.388 & 40.769 & 0.2711 & 34.143 & 2.351 \\
        \midrule
        \midrule
        & DiffBIR & \underline{25.418} & 0.387 & 57.663 & \textbf{0.4131} & \underline{33.074} & \textbf{2.017} \\
        & FlowIE & 24.855 & \underline{0.344} & \underline{50.871} & \underline{0.3954} & 33.390 & 2.208 \\
        & BayesFlow & \textbf{25.731} & \textbf{0.330} & \textbf{40.243} & 0.3440 & \textbf{32.035} & \underline{2.107} \\
        \rowcolor{gray!18} \cellcolor{white} \multirow{-4}{*}{(II)} & LipsFlower (w/ BayesFlow) & 23.409 & 0.508 & 57.037 & 0.1448 & 36.088 & 2.457 \\
        \bottomrule
    \end{tabular}
    \endgroup
\end{table*}

\begin{figure*}[htbp]
    \centering
    \includegraphics[width=\linewidth]
        {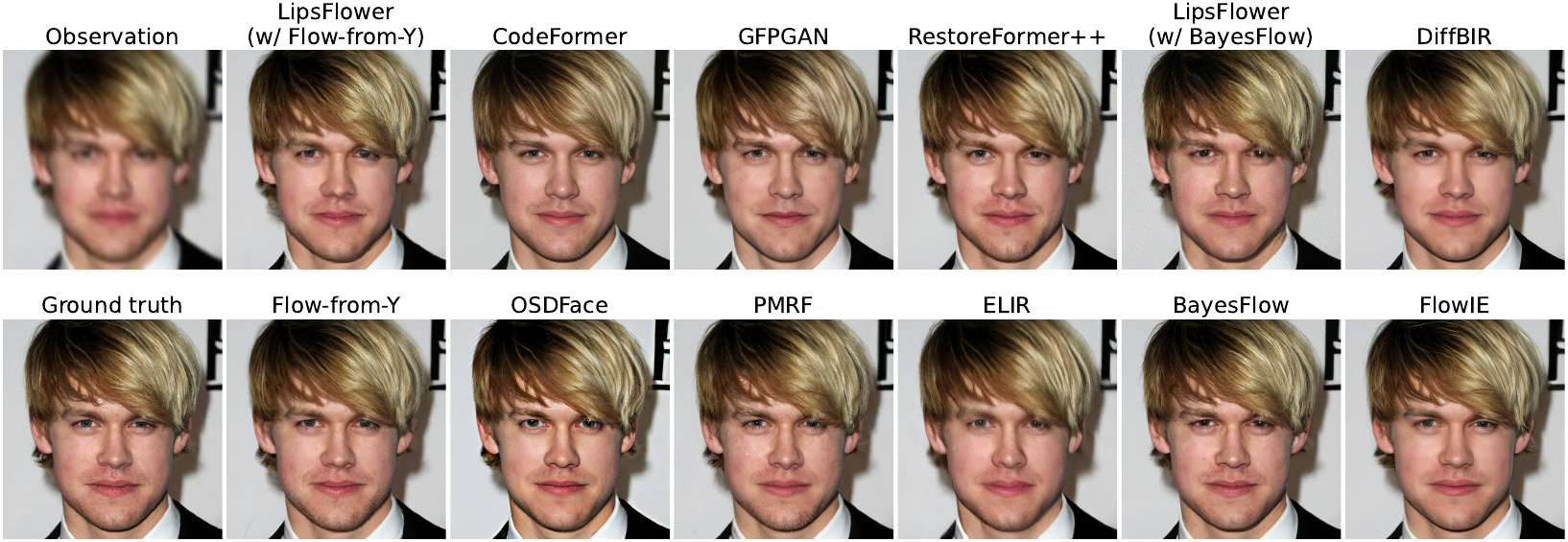}
        \caption{\textbf{Qualitative comparison on CelebA-Test.}
        Both LipsFlower variants closely reproduce the facial structure and hairstyle of the ground truth, with differences mainly in fine texture. Despite the trade-off in perceptual quality for improved robustness, their reconstructions remain visually comparable to those of the other methods.}
    \label{fig:clean_reconstructions}
\end{figure*}

\section{LipsFlower: Architecture, Regularization Details, and Additional Results}
\label{app:lipsflower_details}

This section details the architectural controls and
finite-radius regularization used by LipsFlower.
Both Flow-from-$Y$ and observation-conditioned posterior
sampling use the same core architectural modifications
and localized state regularizer. The conditional model
additionally uses separate input stems and an
observation-sensitivity penalty.

\subsection{LipHDiT Architecture}
\label{app:liphdit_architecture}

\paragraph{Backbone.}
LipHDiT follows the hierarchical organization of
HDiT~\cite{crowson2024scalable}. We use $4\times4$ input
patches, feature widths $(256,512,1024)$, and level depths
$(2,2,8)$. The first two levels use neighborhood attention
with $7\times7$ windows, whereas the coarsest level uses
global attention. The attention head dimension is $64$,
and the feed-forward hidden dimension is three times the
corresponding feature width. The time-conditioning mapping
network has width $768$ and depth $2$. Dropout is disabled.

\paragraph{Stabilized spectral normalization.}
Following spectral normalization~\cite{miyato2018spectral},
we control linear projections using an estimate of the
largest singular value of their weight matrix. For a raw
weight matrix $W$, the effective weight is
\begin{equation}
    \widehat{W}
    =
    \frac{W}{\max(1,\kappa\widehat{\sigma}(W))},
    \qquad \kappa=1.10,
    \label{eq:app_spectral_normalization}
\end{equation}
where $\widehat{\sigma}(W)$ is obtained using power
iteration~\cite{golub2000eigenvalue, miyato2018spectral}. Unlike normalization that always rescales to
unit estimated norm, this rule leaves sufficiently small
weights unchanged.

We use two power iterations per spectral-buffer refresh.
The matrix--vector products and spectral estimates are
computed in single-precision floating-point arithmetic (FP32). To handle zero-initialized projections,
a vanishing iteration product does not replace a valid
iteration vector with a zero vector: the previous valid
direction is retained, with a nonzero fallback when needed.
Consequently, zero initialization does not permanently
disable subsequent spectral estimation.
Spectral-buffer updates are frozen during the paired
evaluations used to compute finite-difference gains,
so both evaluations use the same effective weights.
The factor $\kappa$ provides a practical safety margin,
not a certified upper bound: a finite number of power
iterations can underestimate the true spectral norm.

\paragraph{Normalization and modulation.}
For a token feature vector $h\in\mathbb{R}^{d}$, we use
a denominator-floored variant of
RMSNorm~\cite{zhang2019root}:
\begin{equation}
    \mathcal{N}_{c}(h)
    =
    \frac{h}{
        \max\!\left(
            c,\sqrt{d^{-1}\|h\|_2^2+\varepsilon_{\mathrm{n}}}
        \right)
    },
    \label{eq:app_controlled_rms}
\end{equation}
with $c=0.10$ and $\varepsilon_{\mathrm{n}}=10^{-12}$.
The floor limits the amplification introduced when
normalizing low-magnitude features.

Given the time embedding $e_t$, adaptive normalization
uses scale-only modulation:
\begin{equation}
    \operatorname{AdaNorm}(h,e_t)
    =
    \mathcal{N}_{c}(h)
    \odot
    \left[1+\beta\tanh(Ae_t)\right],
    \label{eq:app_bounded_modulation}
\end{equation}
where $\beta=0.25,$ and $A$ is a spectrally controlled projection.
Thus, each modulation factor lies between $0.75$ and
$1.25$, and no additive modulation shift is used.
This adapts the conditioning mechanisms used in diffusion
transformers~\cite{peebles2023scalable,crowson2024scalable}
while explicitly bounding their multiplicative factors.
Queries and keys are normalized separately using
Eq.~\eqref{eq:app_controlled_rms}, also with floor $0.10$.
This modifies normalized-attention
designs~\cite{henry2020query,crowson2024scalable}
to control their behavior near zero.

\paragraph{Feed-forward transformation.}
The feed-forward expansion projects features into two
equal-sized groups, $a$ and $b$, and applies the
coordinatewise minimum,
\begin{equation}
    \phi(a,b)_i=\min(a_i,b_i).
    \label{eq:app_minout}
\end{equation}
This replaces the multiplicative GEGLU transformation
$a\odot\operatorname{GELU}(b)$~\cite{shazeer2020glu}
with a coordinatewise minimum, followed by a spectrally controlled output projection. The pairwise minimum is nonexpansive under the Euclidean norm on the concatenated input $(a,b)$.

\paragraph{Convex residual and skip mixing.}
Motivated by convex aggregation for Lipschitz
control~\cite{cisse2017parseval}, each attention or
feed-forward residual update takes the form
\begin{equation}
    h^{+}
    =
    (1-\alpha)h+\alpha F_\theta(h),
    \qquad
    \alpha=\alpha_{\max}\operatorname{sigmoid}(a),
    \label{eq:app_convex_residual}
\end{equation}
where $a$ is a learned scalar and $F_\theta$ denotes the
transformation branch alone, excluding the identity skip.
Time conditioning is omitted from the notation.
We set $\alpha_{\max}=0.10$ and initialize $\alpha=0.05$.
The mapping-network residual branches use the same
convex-mixing mechanism.

Decoder token-splitting connections combine the projected,
upsampled features $U(h)$ with the corresponding encoder
features $s$:
\begin{equation}
    h_{\mathrm{dec}}
    =
    (1-\gamma)s+\gamma U(h).
    \label{eq:app_decoder_mixing}
\end{equation}
Here, $\gamma$ is sigmoid-parameterized with maximum
$1$ and initialized to $0.5$.

For an input-independent coefficient $\alpha$ and a
branch with Lipschitz constant $L_F$, the residual map
satisfies
\begin{equation}
    \operatorname{Lip}
    \bigl((1-\alpha)I+\alpha F_\theta\bigr)
    \leq (1-\alpha)+\alpha L_F.
    \label{eq:app_residual_lipschitz}
\end{equation}
This explains the role of convex mixing without implying
that the resulting block is nonexpansive when $L_F>1$.

\paragraph{Conditional input stems.}
The conditional model embeds the state and observation
using separate patch projections:
\begin{equation}
\begin{aligned}
    h_s = E_s(z_t), \quad h_c=E_c(y), \quad
    h_{\mathrm{in}} = P([h_s;\eta h_c]),
\end{aligned}
\label{eq:app_conditional_fusion}
\end{equation}
where $[\cdot;\cdot]$ denotes channel concatenation.
The projections $E_s$, $E_c$, and $P$ use the same
stabilized spectral normalization. The observation
coefficient $\eta$ is sigmoid-parameterized with maximum
$1$ and initialized to $0.75$.
Unlike decoder skip mixing, this module uses scaled
concatenation followed by projection, not a convex sum.

\subsection{Localized Multi-Scale State Perturbations}
\label{app:localized_perturbations}

We adapt adversarial Lipschitz
regularization~\cite{terjek2019adversarial} to sampled
velocity-field inputs. Perturbations are applied only
during the regularizer evaluations, and the inputs and
targets of the main flow-matching loss remain unchanged.

\paragraph{Spatial support.}
For each selected state, we sample a binary rectangular
mask $M\in\{0,1\}^{H\times W}$, shared across channels.
The target area fraction is sampled uniformly from
$[0.02,0.15]$, and the width-to-height aspect ratio is
sampled log-uniformly from $[0.5,2.0]$.
The rectangle dimensions are rounded to valid pixel
dimensions, and its location is sampled uniformly among
valid placements. Thus, the realized area can differ
slightly from the sampled target because of rounding.

\paragraph{Perturbation magnitude.}
Let $C$ be the number of channels and
$n_M=C\sum_{i,j}M_{ij}$ the number of active scalar
coordinates. Define the masked RMS magnitude by
\begin{equation}
    \|u\|_{\mathrm{RMS},M}
    =
    \left(
        \frac{1}{n_M}
        \sum_{c,i,j}M_{ij}u_{cij}^{2}
    \right)^{1/2}.
    \label{eq:app_masked_rms}
\end{equation}
The perturbation radius $r$ is sampled independently
for each selected state according to
Table~\ref{tab:app_radius_mixture}.
All radii are expressed in the coordinates supplied
to the velocity network, rather than in integer
pixel-intensity units.
\begin{table}[t]
    \centering
    \caption{Mixture used to sample the RMS magnitude
    inside the perturbed region.}
    \label{tab:app_radius_mixture}
    \begin{tabular}{lcl}
        \hline
        Scale & Probability & Radius distribution\\
        \hline
        Small & $0.70$ &
        $\operatorname{LogUniform}(0.001,0.01)$\\
        Moderate & $0.20$ &
        $\operatorname{LogUniform}(0.01,0.10)$\\
        Severe & $0.10$ & $r=0.25$\\
        \hline
    \end{tabular}
\end{table}
For an interval $[r_{\min},r_{\max}]$, log-uniform
sampling means
\begin{equation}
    r=
    \exp\!\left(
        \log r_{\min}
        +u\log\frac{r_{\max}}{r_{\min}}
    \right),
    \qquad u\sim\mathcal{U}(0,1).
\end{equation}

\paragraph{Adversarial search.}
Define masked normalization as
\begin{equation}
    \mathcal{Q}_{M}(u)
    =
    \frac{M\odot u}{\|M\odot u\|_{\mathrm{RMS},M}},
    \label{eq:app_masked_normalization}
\end{equation}
with numerical safeguards for zero denominators.
Starting from Gaussian noise $\xi$, we initialize
$\delta^{(0)}=r\mathcal{Q}_{M}(\xi)$.
For the conditional formulation, the search objective is
\begin{equation}
    J_s(\delta)
    =
    \frac{1}{CHW}
    \|v_\theta(z_t+\delta,t;y)
      -v_\theta(z_t,t;y)\|_2^2.
    \label{eq:app_state_attack_objective}
\end{equation}
The observation is omitted for Flow-from-$Y$.

Using projected gradient ascent~\cite{madry2018towards},
we update
\begin{equation}
\begin{aligned}
    q^{(k)} &= \nabla_\delta J_s(\delta^{(k)}),\\
    \delta^{(k+1)}
    &=
    r\,\mathcal{Q}_{M}\!\left(
        \delta^{(k)}
        +\frac{2.5r}{K}\mathcal{Q}_{M}(q^{(k)})
    \right).
\end{aligned}
\label{eq:app_state_attack_update}
\end{equation}
Our configuration uses $K=1$ ascent step.
Each update preserves the sampled support and restores
the prescribed masked RMS magnitude, up to numerical
safeguards. This is a fixed-radius search, rather than
projection onto an RMS ball. No image-range clipping
is applied to the perturbed state.

\paragraph{Gain measurement.}
After the search, the state gain is
\begin{equation}
    g_s
    =
    \frac{
        \|v_\theta(z_t+\delta^{(K)},t;y)
          -v_\theta(z_t,t;y)\|_2
    }{\|\delta^{(K)}\|_2}.
    \label{eq:app_state_gain}
\end{equation}
The implementation evaluates this ratio using full-image
RMS norms, which is equivalent to the Euclidean-norm
ratio for equal input and output dimensions, apart from
numerical safeguards.
The numerator includes every output coordinate and
therefore captures velocity changes outside the
perturbed region.

\subsection{Observation Perturbations for Conditional Sampling}
\label{app:condition_regularization}

For observation-conditioned posterior sampling, we
additionally search for observation perturbations while
holding $z_t$ and $t$ fixed. These perturbations have
whole-image support and prescribed full-image RMS
radius $r_c\in\{0.005,0.01\}$.

The search follows Eq.~\eqref{eq:app_state_attack_update}
with an all-ones mask, radius $r_c$, and objective
\begin{equation}
    J_c(\delta_c)
    =
    \frac{1}{CHW}
    \|v_\theta(z_t,t;y+\delta_c)
      -v_\theta(z_t,t;y)\|_2^2.
\end{equation}
We again use one ascent step, without image-range
clipping. The final observation gain is
\begin{equation}
    g_c
    =
    \frac{
        \|v_\theta(z_t,t;y+\delta_c)-v_\theta(z_t,t;y)\|_2
    }{\|\delta_c\|_2}.
\end{equation}
The two observation radii are cycled during training.
They do not use the localized three-scale mixture
employed by the state regularizer.

\subsection{Time Sampling and Optimization}
\label{app:regularizer_optimization}

\paragraph{Sampling training states.}
We reuse the interpolation states and times generated
for the flow-matching objective~\cite{lipman2023flow}.
Training uses stratified-uniform time sampling.
At each regularization event, two examples per GPU
are selected by cycling through positions in the local
minibatch. This covers different portions of the sampled
training path over successive events, without computing
numerical ODE trajectories.

For the conditional observation search, the radius changes
after a complete cycle through the local minibatch
positions. With the default local batch size of $16$
and two selected examples, each radius is used for
eight regularization events before switching.

\paragraph{Penalty and gradient computation.}
For the selected examples, we compute
\begin{equation}
    \mathcal{P}_s
    =
    \frac{1}{m}\sum_{i=1}^{m}
    [g_{s,i}-\tau_s]_+^2,
    \qquad
    \mathcal{P}_c
    =
    \frac{1}{m}\sum_{i=1}^{m}
    [g_{c,i}-\tau_c]_+^2,
\end{equation}
where $m=2$, $\tau_s=\tau_c=12$, and
$[a]_+=\max(a,0)$.
The observation term is used only for the conditional
model.

During the inner search, the reference velocity is
detached. During the outer optimization, the final
perturbations are detached, while gradients propagate
through both reference and perturbed velocity evaluations.
We therefore do not differentiate through the adversarial
search itself.

\paragraph{Weight warm-up and cadence.}
The nominal weight for each active penalty is
$\lambda_0=10^{-4}$. For zero-based epoch index $e$,
we use
\begin{equation}
    \lambda(e)
    =
    \lambda_0\min\!\left(1,\frac{e+1}{20}\right).
    \label{eq:app_penalty_warmup}
\end{equation}
Thus, the weight increases over the first $20$ epochs, and
regularization is not disabled during this period.

To reduce computation, penalties are evaluated once
every $I$ physical minibatches. At an active event,
their weights are multiplied by $I$ to compensate for
this reduced frequency:
\begin{equation}
    \mathcal{L}_{\mathrm{active}}
    =
    \mathcal{L}_{\mathrm{FM}}
    +I\lambda(e)\mathcal{P}_s
    +I\lambda(e)\mathcal{P}_c,
\end{equation}
with the final term omitted for Flow-from-$Y$.
Inactive minibatches use only the flow-matching loss.

\begin{table}[t]
    \centering
    \caption{Default batching and regularization settings.}
    \label{tab:app_regularizer_settings}
    \begin{tabular}{lcc}
        \hline
        Setting & Flow-from-$Y$ & Conditional\\
        \hline
        Number of GPUs & $4$ & $4$\\
        Batch size per GPU & $8$ & $16$\\
        Gradient accumulation & $8$ & $4$\\
        Effective batch size & $256$ & $256$\\
        Regularization interval $I$ & $8$ & $8$\\
        Selected examples per GPU & $2$ & $2$\\
        Ascent steps per search & $1$ & $1$\\
        Nominal state weight & $10^{-4}$ & $10^{-4}$\\
        Nominal observation weight & --- & $10^{-4}$\\
        Gain threshold & $12$ & $12$\\
        Warm-up epochs & $20$ & $20$\\
        \hline
    \end{tabular}
\end{table}

\subsection{Scope of the Stability Controls}
\label{app:control_scope}

Architectural controls act throughout the network, while
the gain penalty probes selected finite-radius
neighborhoods of training-path states. Neither the
approximate adversarial search nor the power-iteration
estimates provide a global certificate. 
Our propositions therefore remain conditional on their stated
Lipschitz assumptions. The implementation aims to
encourage the sensitivity behavior described by the
analysis. We assess its effect using velocity gains and
end-to-end restoration robustness.

\begin{figure*}[p]
    \centering
    \reconpanel{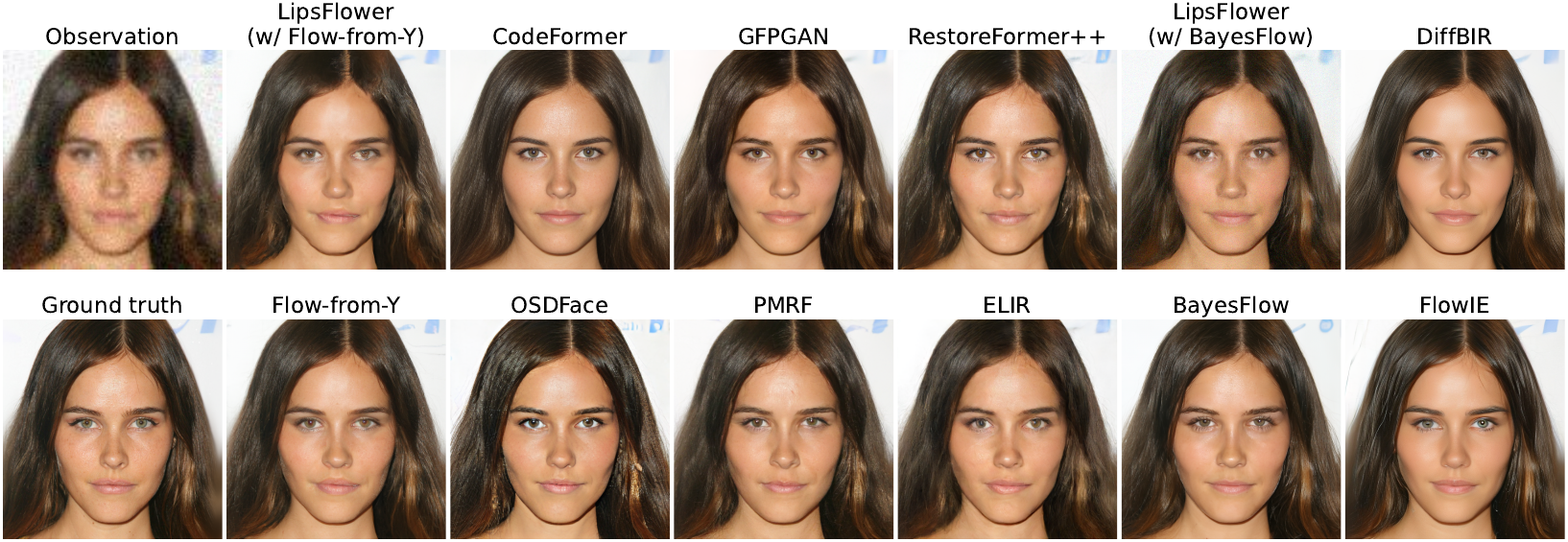}{fig:clean_app_a}
    \reconpanel{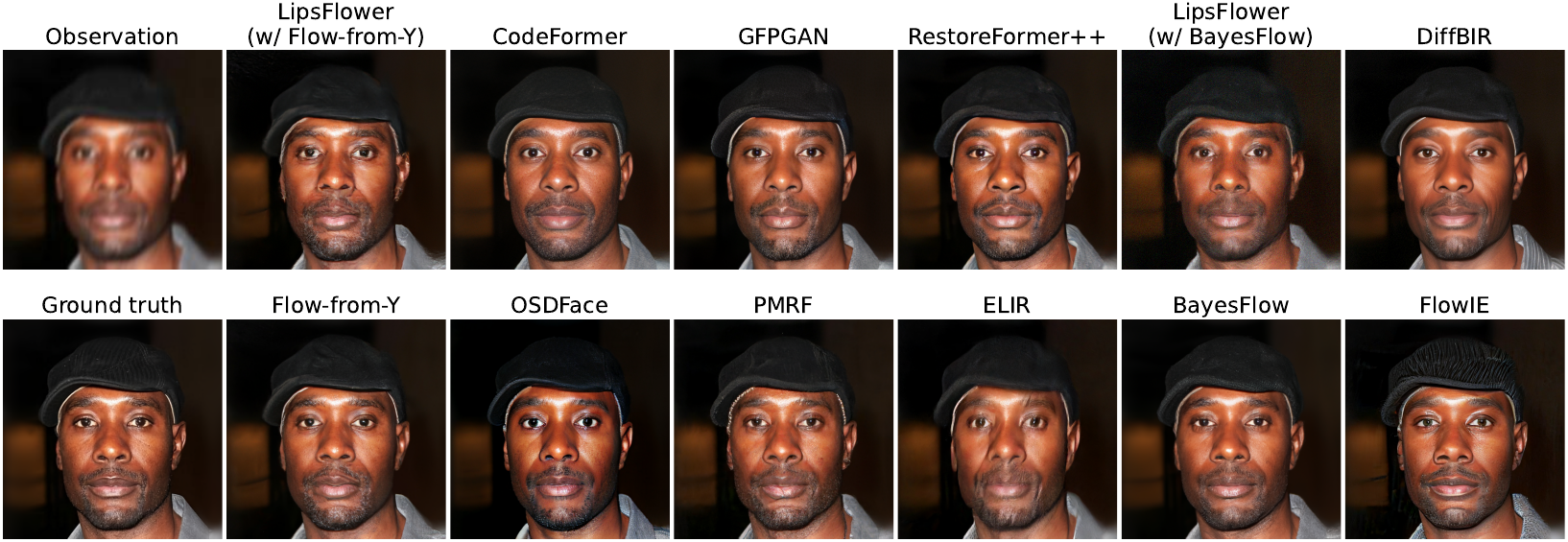}{fig:clean_app_b}
    \reconpanel{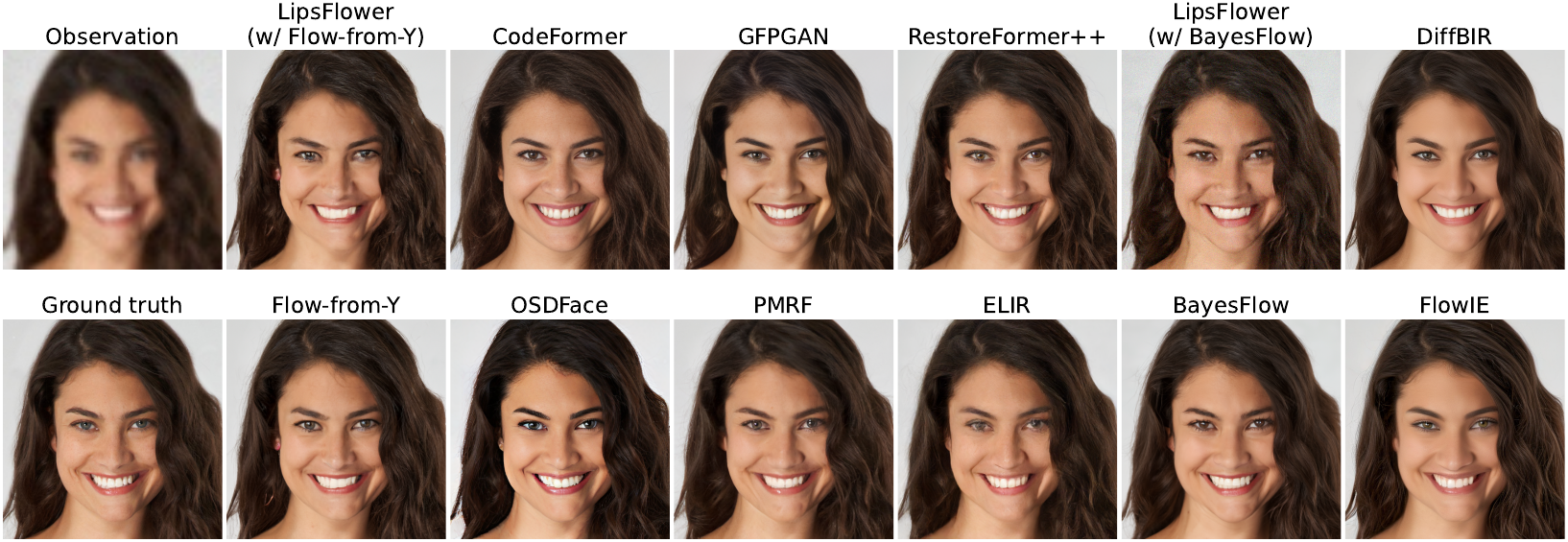}{fig:clean_app_c}
    \caption{\textbf{Additional qualitative comparisons on CelebA-Test.} Panels (a)--(c) show three degraded faces and their clean references alongside reconstructions from both LipsFlower variants and the other methods. The examples illustrate how closely LipsFlower recovers facial structure and finer details such as hair and skin texture.}
    \label{fig:clean_reconstructions_appendix}
\end{figure*}

\subsection{Detailed Restoration Quality Results}
\label{app:restoration_quality}

Greater consistency under input perturbations is useful
only when the model also retains sufficient restoration
quality. Table~\ref{tab:perceptual_quality} therefore
evaluates the same models on unperturbed observations,
measuring reconstruction fidelity, perceptual quality,
and preservation of facial information.

\paragraph{Flow-from-$Y$.}
The robustness improvements come with a relatively
small decrease in PSNR, from $25.784$ to
$25.475$\,dB.
Perceptual metrics also decline: LPIPS increases
from $0.350$ to $0.388$, and FID increases from
$37.050$ to $40.769$.
Nevertheless, the resulting FID remains close to
RestoreFormer++ ($40.611$) and below GFPGAN
($46.160$) and CodeFormer ($52.521$).
The trade-off is therefore quantifiable and does not place our method outside the range of perceptual
quality achieved by the compared restoration methods.
For LipsFlower (w/ Flow-from-$Y$), the central result is lower perturbation amplification with a moderate reduction in perceptual fidelity.

\paragraph{Observation-conditioned sampling.}
The conditional model sacrifices more reconstruction quality.
PSNR decreases from $25.731$ to $23.409$\,dB,
LPIPS increases from $0.330$ to $0.508$, and FID
increases from $40.243$ to $57.037$.
LipsFlower (w/ BayesFlow) thus trades clean reconstruction
quality for lower maximum amplification and perceptual drift.

Figure~\ref{fig:clean_reconstructions} provides a visual perspective on this quality trade-off. In the illustrated example, both LipsFlower variants recover clearly defined facial features from the degraded observation.
The reconstructions retain a coherent facial appearance and overall visual clarity. This example complements the quantitative results by showing that LipsFlower's lower perceptual scores coexist with visually clear reconstructions.

\subsection{Component Ablation Study}
\label{app:component_ablation}

\begin{table*}[t]
    \centering
      \caption{\textbf{Component ablation for Flow-from-$Y$ under SPSA attacks on CelebA-Test.} A and B use standard HDiT, without and with sensitivity regularization, respectively; C uses LipHDiT without regularization; E combines both components (full LipsFlower). Lower is better: bold and underlined values indicate the best and second-best point estimates, respectively.}
    \label{tab:spsa_ablation}
    \begingroup
    \scriptsize
    \setlength{\tabcolsep}{2.5pt}
    \renewcommand{\arraystretch}{1.12}
    \begin{tabular}{llc*{10}{r}}
        \toprule
        \multirow{2}{*}{Setting} & \multirow{2}{*}{Backbone}
        & \multirow{2}{*}{Gain reg.}
        & \multicolumn{5}{c}{SPSA--$\ell_2$ (RMS budget)}
        & \multicolumn{5}{c}{SPSA--$\ell_\infty$} \\
        \cmidrule(lr){4-8}\cmidrule(lr){9-13}
        & & & Mean $G$ & P95 $G$ & $\mathrm{CVaR}_{95}\,G$ & Max $G$ & LPIPS drift
        & Mean $G$ & P95 $G$ & $\mathrm{CVaR}_{95}\,G$ & Max $G$ & LPIPS drift \\
        \midrule
        A: Baseline & HDiT & --
        & 0.525 & 0.743 & 0.849 & 1.740 & 15.803
        & 0.523 & 0.737 & 0.830 & 1.660 & 9.188 \\
        B: Regularizer only & HDiT & Yes
        & 0.559 & 0.791 & 0.904 & 1.584 & 14.040
        & 0.558 & 0.770 & 0.878 & 1.631 & 8.660 \\
        C: Architecture only & LipHDiT & --
        & \underline{0.486} & \underline{0.688} & \underline{0.775} & \underline{1.309} & \underline{12.530}
        & \underline{0.489} & \underline{0.674} & \underline{0.750} & \underline{1.277} & \underline{7.772} \\
        \rowcolor{gray!18}
        E: LipsFlower & LipHDiT & Yes
        & \textbf{0.475} & \textbf{0.667} & \textbf{0.747} & \textbf{1.158} & \textbf{11.649}
        & \textbf{0.480} & \textbf{0.658} & \textbf{0.727} & \textbf{1.118} & \textbf{7.360} \\
        \bottomrule
    \end{tabular}
    \endgroup
\end{table*}

\begin{table*}[t]
    \centering
    \caption{\textbf{Clean reconstruction and perceptual quality for the Flow-from-$Y$ component ablation on CelebA-Test.} A and B use standard HDiT, without and with sensitivity regularization, respectively; C uses LipHDiT without regularization; E combines both components (full LipsFlower). Bold and underlined values indicate the best and second-best point estimates, respectively.}
    \label{tab:perceptual_quality_ablation}
    \begingroup
    \small
    \setlength{\tabcolsep}{3pt}
    \renewcommand{\arraystretch}{1.12}
    \begin{tabular}{llcrrrrrr}
        \toprule
        Setting & Backbone & Gain reg. & PSNR $\uparrow$ & LPIPS $\downarrow$
        & FID $\downarrow$ & Precision $\uparrow$
        & DEG $\downarrow$ & LMD $\downarrow$ \\
        \midrule
        A: Baseline & HDiT & --
        & \textbf{25.784} & \textbf{0.350} & \underline{37.050} & \underline{0.3797}
        & \textbf{30.602} & \textbf{1.995} \\
        B: Regularizer only & HDiT & Yes
        & 25.298 & 0.384 & \textbf{35.306} & \textbf{0.4005}
        & \underline{31.145} & \underline{2.045} \\
        C: Architecture only & LipHDiT & --
        & 25.474 & \underline{0.379} & 39.906 & 0.2625
        & 34.148 & 2.345 \\
        \rowcolor{gray!18}
        E: LipsFlower & LipHDiT & Yes
        & \underline{25.475} & 0.388 & 40.769 & 0.2711
        & 34.143 & 2.351 \\
        \bottomrule
    \end{tabular}
    \endgroup
\end{table*}

\begin{table*}[ht]
    \centering
    \caption{Inference configurations for the literature
    baselines. Iterative methods retain their respective
    sampling procedures.}
    \label{tab:baseline_inference}
    \small
    \setlength{\tabcolsep}{5pt}
    \renewcommand{\arraystretch}{1.15}
    \begin{threeparttable}
    \begin{tabular}{p{0.17\textwidth}p{0.76\textwidth}}
        \toprule
        Method & Inference configuration\\
        \midrule
        CodeFormer~\tnote{(a)} &
        Fidelity weight $0.5$; aligned inputs;
        upscale factor $1$; background upsampling
        disabled.\\

        GFPGAN~\tnote{(b)}  &
        Released v1.3 checkpoint with the clean
        architecture; restoration weight $0.5$;
        aligned inputs; upscale factor $1$.\\

        RestoreFormer++~\tnote{(c)} &
        Released RestoreFormer++ checkpoint;
        aligned inputs; upscale factor $1$.\\

        OSDFace~\tnote{(d)} &
        Stable Diffusion 2.1-base backbone;
        one U-Net evaluation at timestep $399$;
        merged LoRA with rank $16$ and scale
        parameter $16$; FP32 inference.\\

        DiffBIR~\tnote{(e)} &
        Version 2 face-restoration pipeline;
        spaced sampler with $50$ steps;
        classifier-free guidance scale $4$;
        captioning disabled; FP32 inference.\\

        PMRF~\tnote{(f)} &
        Official implementation and pretrained checkpoint; $100$ flow integration steps.\\

        FlowIE~\tnote{(g)} &
        Released blind-face-restoration checkpoint;
        official mean-value sampler with midpoint
        index $3$, requiring four velocity evaluations;
        preprocessing enabled and wavelet color
        correction.\\

        ELIR~\tnote{(h)} &
        Released blind-face-restoration checkpoint;
        five flow steps; source-noise scale $0.1$;
        latent dimensions $16\times64\times64$.\\
        \bottomrule
    \end{tabular}
    \begin{tablenotes}
        \item[a] \url{https://github.com/sczhou/CodeFormer}
        \item[b] \url{https://github.com/TencentARC/GFPGAN} 
        \item[c] \url{https://github.com/wzhouxiff/RestoreFormerPlusPlus} 
        \item[d] \url{https://github.com/jkwang28/OSDFace}
        \item[e] \url{https://github.com/XPixelGroup/DiffBIR}
        \item[f] \url{https://github.com/ohayonguy/PMRF}
        \item[g] \url{https://github.com/EternalEvan/FlowIE}
        \item[h] \url{https://github.com/eladc-git/ELIR}
    \end{tablenotes}
    \end{threeparttable}
\end{table*}

Tables~\ref{tab:spsa_ablation}
and~\ref{tab:perceptual_quality_ablation} examine
the contributions of architectural controls and
gain regularization in Flow-from-$Y$.
We compare four configurations: standard HDiT (A),
HDiT with gain regularization (B), LipHDiT without
gain regularization (C), and full LipsFlower (E).
All configurations use $1{,}000$ training epochs and $100$ Euler sampling steps.

\paragraph{Architectural controls.}
Replacing HDiT with LipHDiT without adding the
regularizer (A$\rightarrow$C) improves every reported
robustness measure under both attack constraints.
For example, under the RMS-constrained attack,
mean gain decreases from $0.525$ to $0.486$,
and the average gain among the worst $5\%$ of images
decreases from $0.849$ to $0.775$.
Maximum gain decreases from $1.740$ to $1.309$.
These results show that the architectural controls
provide a substantial part of the observed
robustness improvement.

\paragraph{Gain regularization.}
The effect of the regularizer depends on the backbone.
Adding it to standard HDiT (A$\rightarrow$B)
reduces maximum gain and perceptual drift, but
increases mean gain, the 95th percentile, and
$\mathrm{CVaR}_{95}$ under both attacks.
Thus, regularization alone does not consistently
reduce end-to-end sensitivity.

Adding the same regularizer to LipHDiT
(C$\rightarrow$E), however, improves every reported
robustness measure. Maximum gain decreases from
$1.309$ to $1.158$ under the RMS-constrained attack
and from $1.277$ to $1.118$ under the
$\ell_\infty$ attack, reductions of approximately
$11.5\%$ and $12.5\%$.
Full LipsFlower consequently achieves the lowest
observed sensitivity across all reported measures
among the four configurations.

\paragraph{Restoration-quality trade-off.}
The quality results clarify the cost of these
improvements. Architecture-only LipHDiT and full
LipsFlower have nearly identical PSNR
($25.474$ and $25.475$\,dB).
Adding regularization nevertheless increases LPIPS
from $0.379$ to $0.388$ and FID from $39.906$
to $40.769$, while slightly improving generative
precision.
The additional robustness therefore comes with
some loss of perceptual quality despite essentially
unchanged PSNR.

Quality changes are also metric-dependent.
Regularizer-only HDiT achieves the best FID
($35.306$) and precision ($0.4005$), whereas the
unregularized baseline achieves the best PSNR,
LPIPS, identity similarity, and landmark accuracy.
Taken together, the results suggest that architectural controls and gain regularization work best in combination for measured robustness, with more varied effects on perceptual quality.

\section{Training and Inference Settings}
\label{app:training_inference}

This section describes the training configuration of
LipsFlower and the inference procedures used for the
main comparisons. 

\subsection{Training Data and Degradation Pipeline}
\label{app:training_data}

We train our restoration models on FFHQ images~\cite{karras2019style} at
$512\times512$ resolution. Images are converted to
three-channel RGB tensors with values in $[0,1]$.
Degraded observations are generated by combining blur, resizing, Gaussian noise, and JPEG compression.
The pipeline samples isotropic or anisotropic Gaussian
blur with equal probability, using a $41\times41$
kernel and blur standard deviations in $[0.1,15]$.
The resizing factor is sampled from $[0.8,32]$,
the Gaussian noise standard deviation from $[0,20/255]$,
and JPEG quality from $[30,100]$.
The degraded image is subsequently resized back to
$512\times512$ using bilinear interpolation.
Training also uses random horizontal flipping and
grayscale conversion with probability $0.01$,
while retaining three channels.
The resulting observation is quantized to
$1/255$ increments.
Consequently, the clean image and degraded observation
are supplied in the same spatial and channel dimensions.
Downsampling and resizing back to the target resolution are part of observation generation, and the restoration model does not receive the sampled degradation parameters.

\subsection{Optimization}
\label{app:training_optimization}

Both LipsFlower formulations use AdamW~\cite{loshchilov2017decoupled} with a constant
learning rate of $5\times10^{-4}$,
$(\beta_1,\beta_2)=(0.9,0.95)$, and optimizer
$\epsilon=10^{-8}$.
We clip the gradient norm to $1$ and maintain
an exponential moving average (EMA) of the model
parameters with decay $0.9999$.
Weight decay is $10^{-2}$ for ordinary parameters
and disabled for the raw weights of spectrally
controlled projections.
All optimizer groups use the same learning rate.
Table~\ref{tab:training_configuration} summarizes the
reference batching configuration.
Gradient accumulation gives an effective batch size
of $256$ for both formulations.

\begin{table}[H]
    \centering
    \caption{Reference training configuration for
    the two LipsFlower formulations. Batch size
    denotes the number of examples per GPU.}
    \label{tab:training_configuration}
    \small
    \begin{tabular}{lcc}
        \toprule
        Setting & Flow-from-$Y$ & Conditional\\
        \midrule
        Image resolution & $512^2$ & $512^2$\\
        Number of GPUs & $4$ & $4$\\
        Batch size per GPU & $8$ & $16$\\
        Accumulation steps & $8$ & $4$\\
        Effective batch size & $256$ & $256$\\
        Learning rate & $5\times10^{-4}$ & $5\times10^{-4}$\\
        EMA decay & $0.9999$ & $0.9999$\\
        Gradient clipping norm & $1$ & $1$\\
        \bottomrule
    \end{tabular}
\end{table}

For Flow-from-$Y$, the source state is $z_0=y+0.1\epsilon$, where $\epsilon\sim\mathcal{N}(0,I)$. Training uses linear interpolation between this source and the clean target, with the displacement $x-z_0$ as the velocity-regression target. By contrast, conditional sampling uses independently sampled standard Gaussian noise as the source, with the observation supplied as a separate condition. 

\subsection{Inference for the Flow-Matching Comparisons}
\label{app:flow_inference}

The main evaluation uses $100$ forward-Euler steps and EMA parameters for both Flow-from-$Y$ models and both conditional models. Training-only gain penalties are disabled at inference. Both inference pipelines disable automatic mixed
precision. Final reconstructions are clipped to $[0,1]$ and saved as eight-bit RGB images.

\paragraph{Flow-from-$Y$.}
Saved degraded observations are passed directly to
the model without additional degradation.
Each image receives a Gaussian source-noise draw
generated from the evaluation seed and its position
in the input sequence.
The input order and seed (42) are held fixed across
unperturbed and perturbed evaluations, ensuring
that each pair uses the same source noise.
After loading EMA parameters, spectral buffers are
refreshed using at least $10$ power iterations.

\paragraph{Conditional sampling.}
We restore the saved observation-standardization
statistics before loading the final EMA velocity
parameters.
Each observation produces one conditional posterior
sample.
The evaluation uses seed $42$ and a fixed Gaussian
source draw across observations and perturbation
conditions.
For the conditional LipsFlower model, spectral
buffers are refreshed using five power iterations
after loading EMA parameters.

\subsection{Inference for Literature Baselines}
\label{app:baseline_inference}

We use released pretrained weights for the literature
baselines and retain their method-specific inference
procedures.
Table~\ref{tab:baseline_inference} lists the settings
configured by the evaluation adapters.
Thus, the comparison uses a common test cohort and
attack protocol, but does not impose an identical
number of network evaluations on every method.

\subsection{Shared Evaluation Procedure}
\label{app:shared_evaluation}

All methods are evaluated on the same $3{,}000$
paired CelebA-Test observations and ground truths.
The evaluation seed is $42$.
For each method, inference settings and random-seed
handling are retained across clean and attacked
inputs to separate observation sensitivity from
sampling variability.
Repeated inference on unperturbed inputs provides
a repeatability check.
The quality and robustness tables use reconstructions
from the same evaluation cohort.
Metrics are computed from saved eight-bit images;
gain denominators use the realized input differences
after clipping and quantization.
FID and generative precision use the same FFHQ
reference set for every method.

\begin{figure*}[p]
    \centering
    \reconpanel{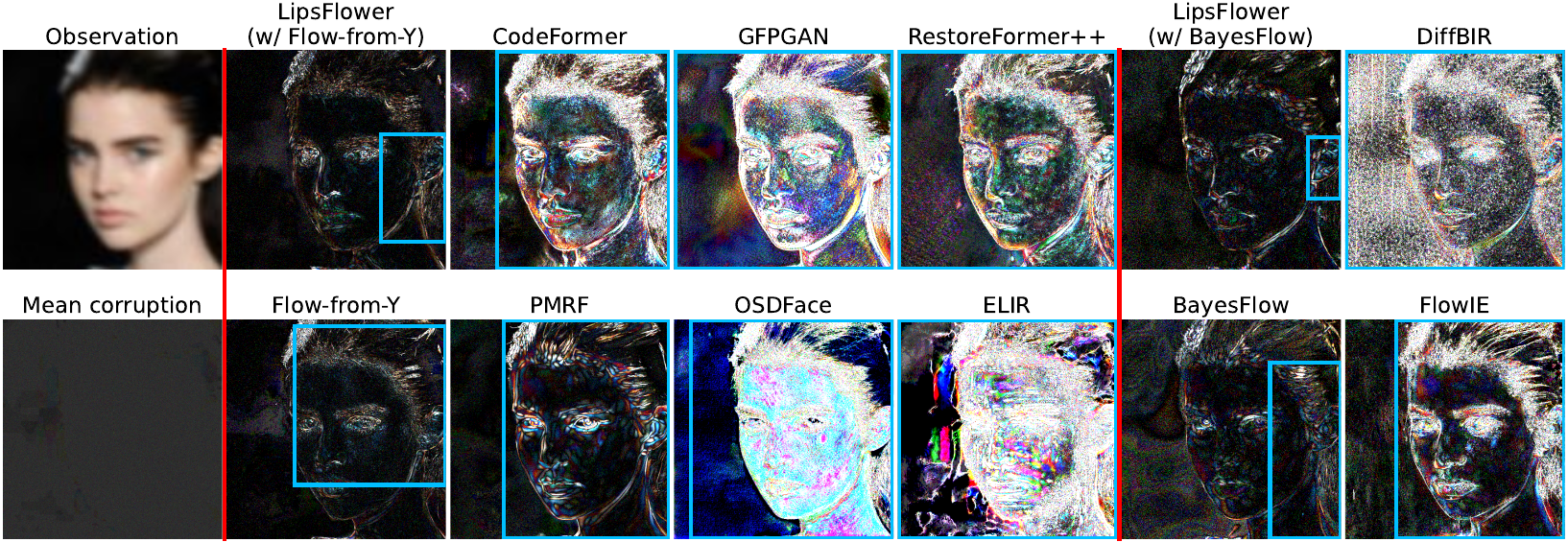}{fig:change_app_a}
    \reconpanel{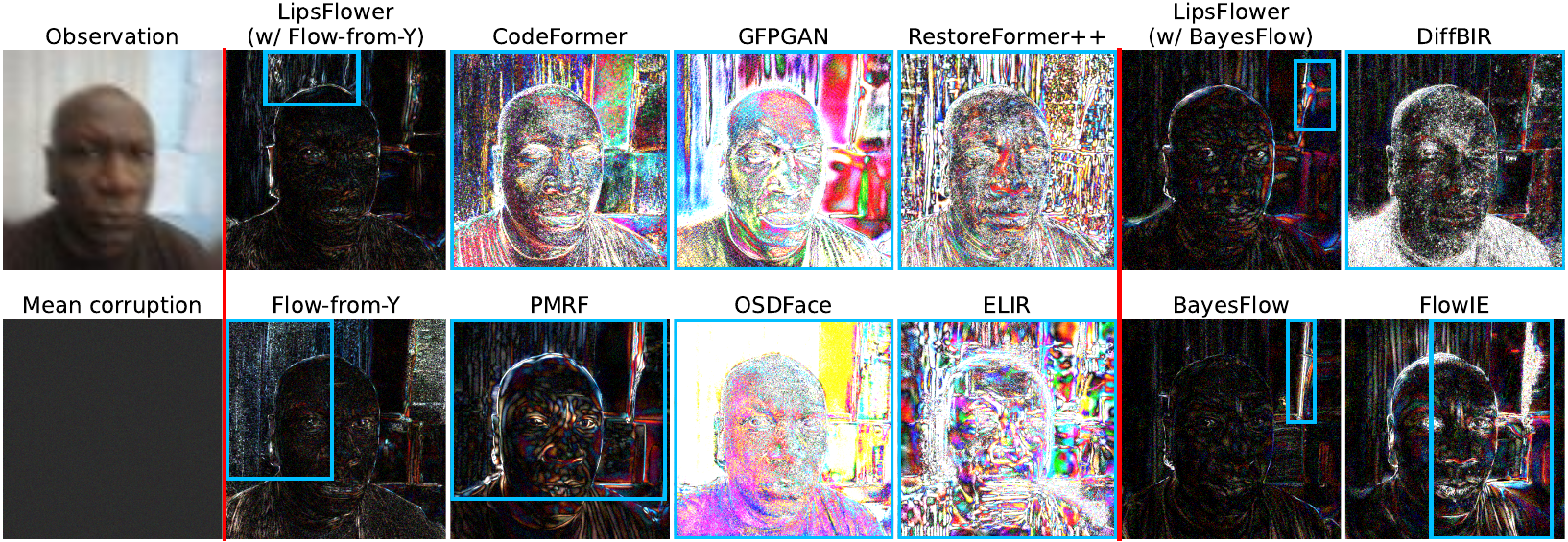}{fig:change_app_b}
    \reconpanel{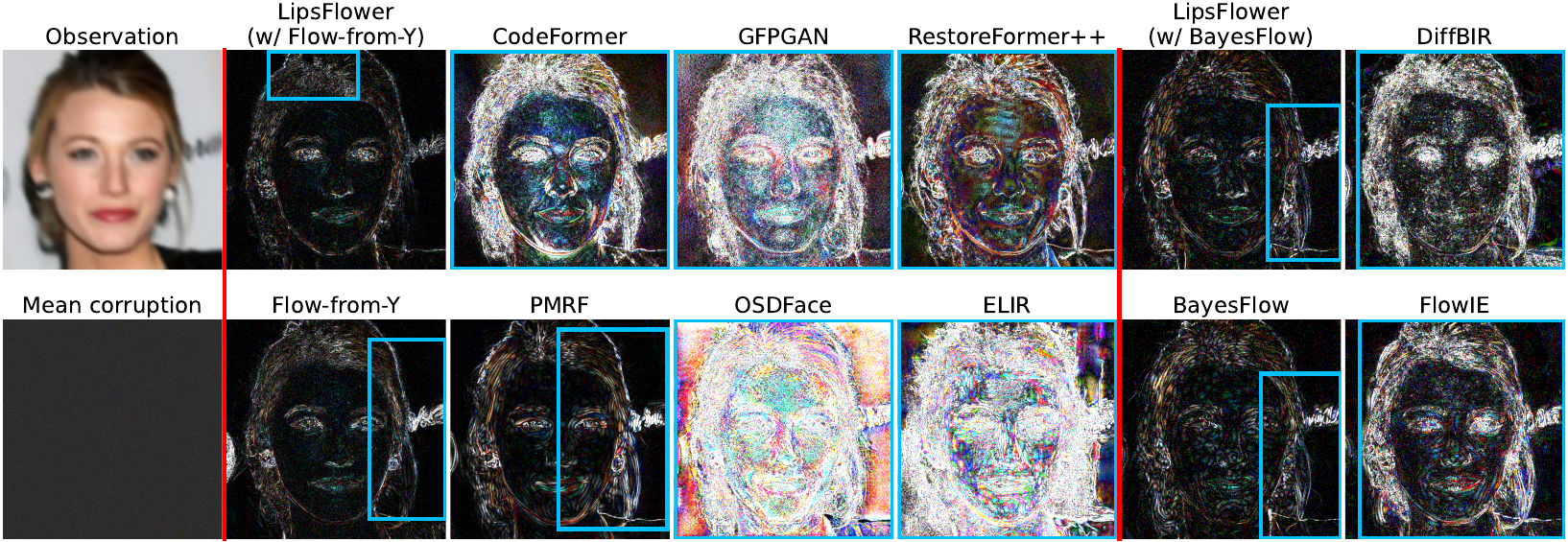}{fig:change_app_c}
    
    \caption{\textbf{Output changes under adversarial perturbations on CelebA-Test.}
    For three degraded faces, panels (a)--(c) show the amplified absolute difference between each method's reconstructions before and after a model-specific black-box SPSA attack on the observation. Consistent with the quantitative results, LipsFlower reduces output changes for both backbones, with a larger reduction for Flow-from-$Y$.}
    \label{fig:output_change_appendix}
\end{figure*}


\end{document}